\documentclass[runningheads]{llncs}
 
\usepackage{eccv}

\usepackage{eccvabbrv}
\usepackage{graphicx}
\usepackage{booktabs}
\usepackage{multirow}
\usepackage{pifont}
\usepackage{booktabs}
\usepackage{tabularx}
\usepackage{makecell}
\usepackage{multirow}
\usepackage{xcolor}
\usepackage{colortbl}
\usepackage{enumitem}
\usepackage{wrapfig}
\usepackage{soul}
\definecolor{groupgray}{RGB}{245,245,245}
\definecolor{canonblue}{RGB}{25,55,95}
\definecolor{syngray}{RGB}{140,140,140}
\newcommand{\cmark}{\textcolor{green!70!black}{\ding{51}}}
\newcommand{\xmark}{\textcolor{red!70!black}{\ding{55}}}

\usepackage{xcolor}
\usepackage{colortbl}
\usepackage{enumitem}
\definecolor{groupgray}{RGB}{245,245,245}
\definecolor{canonblue}{RGB}{25,55,95}
\definecolor{syngray}{RGB}{140,140,140}
\usepackage[skins]{tcolorbox}
\tcbuselibrary{listings,breakable}
\usepackage{caption}
\newtcolorbox{systemprompt}{
    colback=gray!5,
    colframe=gray!60,
    boxrule=0.5pt,
    arc=3pt,
    left=3pt,
    right=3pt,
    top=3pt,
    bottom=3pt,
    listing only,
    listing options={
        basicstyle=\ttfamily\fontsize{5}{5.5}\selectfont,
        breaklines=true,
        columns=fullflexible
    }
}
\newtcolorbox{promptbox}{
  colback=gray!10,    
  colframe=gray!60,  
  arc=1mm,           
  boxrule=0.8pt,     
  left=3pt, right=3pt, top=3pt, bottom=3pt,
    fontupper=\small, 
  enhanced
}
\usepackage[accsupp]{axessibility}  

\usepackage{hyperref}

\usepackage{orcidlink}

\begin{document}
\title{Natural Language Camera Movement Understanding} 

\titlerunning{Natural Language Camera Movement Understanding}

\author{
Yuwen Tan\inst{1}\orcidlink{0009-0004-2417-2955} \and
Joey Huang\inst{1}\orcidlink{0009-0009-9427-1955} \and
Jin Huang\inst{2}\orcidlink{0000-0002-1803-7077} \and
Haoxiang Li\inst{2}\orcidlink{0009-0006-1525-3942} \and
Boqing Gong\inst{1}\orcidlink{0000-0003-3915-5977}
}

\authorrunning{Y.\ Tan et al.}

\institute{Boston University, Boston, MA 02215, USA 
\and
Pixocial Technology, Bellevue, WA 98004, USA 
\\[4mm]
\small{\url{https://1yuwen.github.io/ACaM-Project-Page/}}}
\maketitle

\begin{abstract}
Understanding camera movement in natural language is critical for training and evaluating video generation models, among other applications. However, we demonstrate that existing vision-language models (VLMs) fail this task in surprising ways, frequently confusing translation with rotation, left with right, and object movement with camera movement. To address these limitations, we establish natural language camera movement understanding as a standalone research task. We introduce a two-level cinematographic taxonomy and an extensive, atomic benchmark featuring both real and synthetic videos. Furthermore, we curate a large-scale, multi-source training set enhanced by targeted camera movement augmentation. Our fine-tuned VLM-8B outperforms {Gemini 3.1 Pro} by {10}\% and {11}\% on our benchmark's real and synthetic videos, respectively. Despite these gains, a significant gap remains relative to human performance, underscoring the need to promote and facilitate future research on natural language camera movement understanding.

\end{abstract}


\section{Introduction}
\label{sec:intro}

\begin{figure}[t]
  \centering   \includegraphics[width=0.99\linewidth]{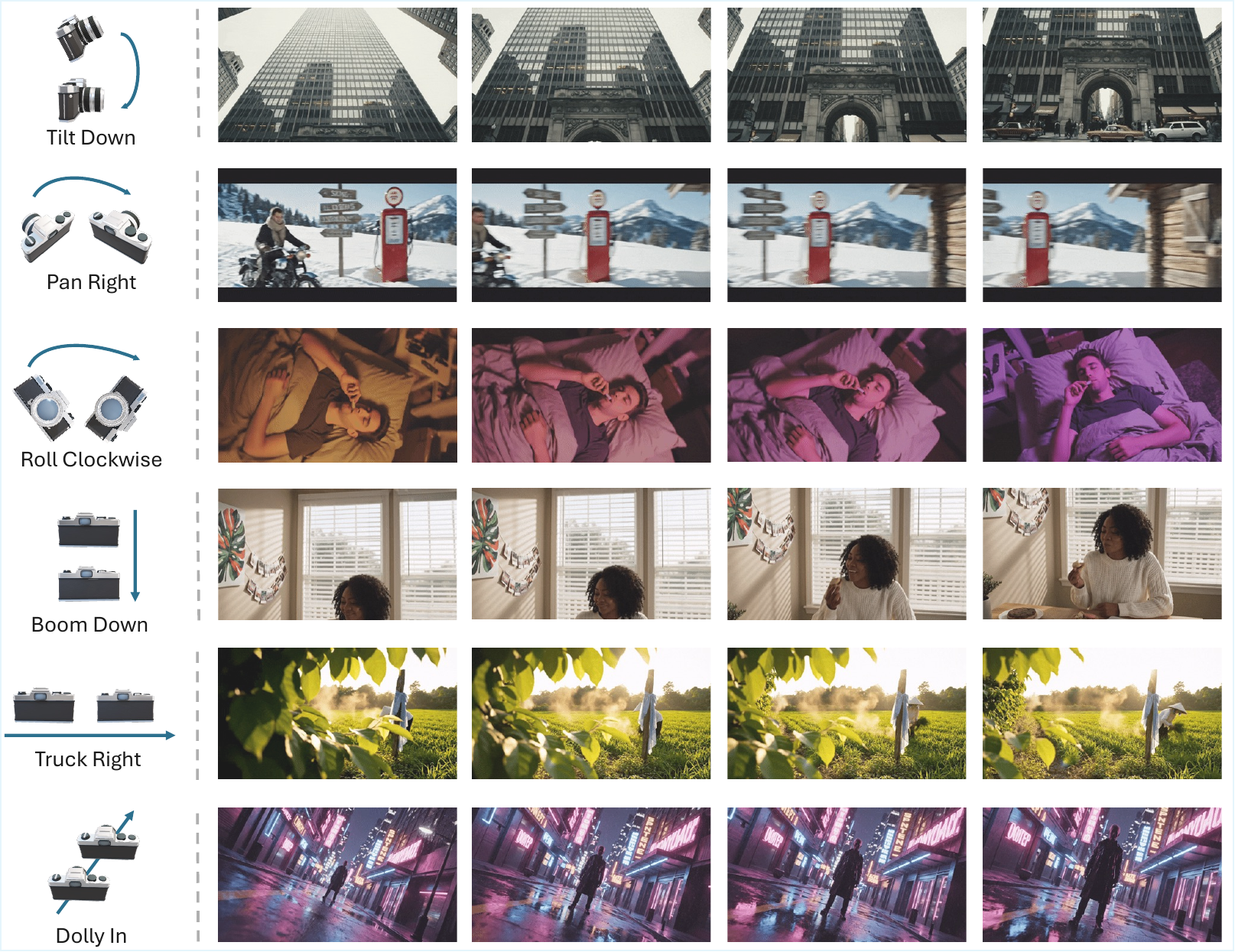}
  \caption{Some examples of camera movements in cinematographic practice: Camera rotations (pan, tilt, roll) and translations (boom, dolly, truck).}
\label{fig:camera_movement_example}
\end{figure}

Camera movement is one of the most powerful tools in filmmaking~\cite{nielsen2007camera,keating2019dynamic,bordwell2008film,brown2016cinematography,spottiswoode1969grammar}, influencing how audiences perceive and engage with the narrative. Filmmakers deliberately use specific camera movements to serve distinct storytelling purposes. For instance, a `dolly in' shot can emphasize a character's internal realization, while a pan shot can reveal new spatial details (see more examples of camera movement in Fig.~\ref{fig:camera_movement_example}). 
In recent years, modern video generation models~\cite{wan2025wan,sora,veo,Hailuo,Kling,kong2024hunyuanvideo,yang2024cogvideox} have begun to explicitly incorporate camera movement into the generation process; users are now describing camera movement using cinematographic text prompts (e.g., `A slow pan') to guide video generation.

Natural language understanding of camera movement is more important than ever. As modern text-to-video generation models are increasingly being used to generate dynamic scenes with varied camera movements, enhancing the controllability of these specific movements heavily depends on massive, high-quality video-text pairs. Therefore, robust models capable of automatically filtering, curating, and captioning cinematographic camera movements in large-scale, unannotated videos are becoming increasingly relied upon to eliminate the bottleneck of expensive manual annotation. Simultaneously, in terms of model evaluation~\cite{mou2025gradeo,wang2025love,jia2025vqa2,sun2025t2v}, the ability to automatically and reliably verify whether generated videos faithfully execute specific camera prompts has become indispensable for benchmarking modern generative systems.

 Unlike traditional camera pose estimation tasks~\cite{CUTR3,wang2025vggt,vipe,megasam,pan2024global}, which estimate per-frame poses in coordinate space, natural language understanding operates at the semantic level. It maps spatio-temporal visual cues directly to human-interpretable cinematographic languages. While geometry-based methods provide accurate trajectories, they are not as accessible to common users as language prompts and often fail to infer object-centric camera movements~\cite{camerabench,camreasoner}, such as `tracking' and `arc'. Furthermore, their reliance on depth estimation and calibrated intrinsics makes them computationally heavy and unnecessarily complex for natural language camera movement understanding. We believe vision-language models (VLMs)~\cite{llavaov,bai2025qwen2,gpt4o,gemini,internvl3,internvideo2.5,bai2025qwen3,internvl3.5} offer a more suitable and scalable framework for this task.

Drawing upon established cinematography literature~\cite{nielsen2007camera,keating2019dynamic}, we categorize various camera movements into a two-level taxonomy of 17 distinct classes (summarized in Tab.~\ref{tab:motion_taxonomy}). This taxonomy comprehensively covers camera translations, rotations, focal-length changes, object-centric movements, and static shots. Crucially, it provides a channel to systematically investigate the capability of camera movement understanding in VLMs.  Although recent works~\cite{shotbench,camerabench,cinetechbench,li2024can,wu2025refineshot,motionsight} have observed that VLMs struggle significantly with this task, the underlying reasons for their poor performance remain largely unexplored. In this work, we conduct an in-depth analysis to investigate \textit{why} camera movement understanding is so challenging for VLMs. Through preliminary evaluations, we find that their inherent deficiencies fall under five failure modes:{
 VLMs are insensitive to subtle inter-frame changes, confuse physical movement with angular change, and frequently misinterpret movement direction. Furthermore, they struggle to distinguish between optical zoom and physical dolly, and also conflate object movement with global camera movement.
}



\begin{table}[t]
\centering
\caption{Taxonomy of camera movements with cinematographic terminology.}
\label{tab:motion_taxonomy}
\setlength{\tabcolsep}{4.5pt}
\scriptsize
\begin{tabular}{@{} l p{0.30\linewidth} p{0.30\linewidth} @{}}
\toprule
\multirow{8}{*}{\textbf{Translation}}
& \textcolor{canonblue}{\textbf{Dolly In}} \newline {\scriptsize \color{syngray} (Move in, Push in, Move forward)}
& \textcolor{canonblue}{\textbf{Dolly Out}} \newline {\scriptsize \color{syngray} (Move out, Pull out, Move backward)} \\
& \textcolor{canonblue}{\textbf{Truck Left}} \newline {\scriptsize \color{syngray} (Move to the left)}
& \textcolor{canonblue}{\textbf{Truck Right}} \newline {\scriptsize \color{syngray} (Move to the right)} \\

& \textcolor{canonblue}{\textbf{Boom Up}} \newline {\scriptsize \color{syngray} (Pedestal up, Crane up, Move upwards)}
& \textcolor{canonblue}{\textbf{Boom Down}} \newline {\scriptsize \color{syngray} (Pedestal down, Crane down, Move downwards)} \\

\midrule

\multirow{3}{*}{\textbf{Rotation}}
& \textcolor{canonblue}{\textbf{Pan Left}}
& \textcolor{canonblue}{\textbf{Pan Right}} \\

& \textcolor{canonblue}{\textbf{Tilt Up}}
& \textcolor{canonblue}{\textbf{Tilt Down}} \\

& \textcolor{canonblue}{\textbf{Roll Counterclockwise}}
& \textcolor{canonblue}{\textbf{Roll Clockwise}} \\

\midrule

\textbf{Focal Length Change}
& \textcolor{canonblue}{\textbf{Zoom In}}
& \textcolor{canonblue}{\textbf{Zoom Out}} \\

\midrule
\textbf{Static}
& \multicolumn{2}{l}{\textcolor{canonblue}{\textbf{Static}} \quad {\scriptsize \color{syngray} (Stationary, Fixed, No movement)}} \\

\midrule

\rowcolor{groupgray}
\textbf{Object-centric Movement}
& \textcolor{canonblue}{\textbf{Tracking}} \newline {\scriptsize \color{syngray} (Follow shot)}
& \textcolor{canonblue}{\textbf{Arc}} \newline {\scriptsize \color{syngray} (Orbit)} \\
\bottomrule
\end{tabular}
\end{table}


These challenges are nontrivial. We tackle them by first promoting natural language camera movement understanding as a standalone research problem rather than allowing it to become a minor part of fine-grained motion~\cite{favorbench,motionbench,motionsight}, cinematography~\cite{cinetechbench,shotbench,movienet,cinescale2,wu2025refineshot,rao2020unified,li2024can,tang2025vidcomposition}, and camera motion primitives~\cite{camerabench}. We then instantiate it with a comprehensive and atomic benchmark containing real-world cinematography and synthetically generated videos, in response to the needs of curating real-world video-text training data and automatic rating of modern video generation models. {We purposely choose atomic camera movements in the benchmark---one target camera movement per video clip for evaluation---leaving the combination of multiple movements for future work.} Finally, we strive to deliver the best-performing VLM model for camera movement understanding by constructing a large-scale training dataset annotated with camera movement labels from five data sources and applying targeted data augmentation to explicitly enhance motion perception. We fine-tune Qwen3-VL-4B and 8B~\cite{bai2025qwen3} on the curated dataset, 
achieving relative improvements of 10\% on real-world videos and {11}\% on synthetic videos 
over the strong proprietary model {Gemini-3.1-Pro}. However, a human study still exposes a substantial gap in performance between our best VLM and human, thus calling for future and more extensive research on natural language camera movement understanding. 

\section{Why is camera movement understanding challenging for current VLMs?}
\label{sec:why_hard}
To provide an initial understanding of why this task is so challenging for VLMs, we conduct preliminary evaluations and present several illustrative examples. We find that their deficiencies fall under five failure modes. We elaborate on each of these failure modes in the following subsections. Detailed experimental setups are provided in the supplementary material.

\begin{figure}[tb]
  \centering
\includegraphics[width=0.99\linewidth]{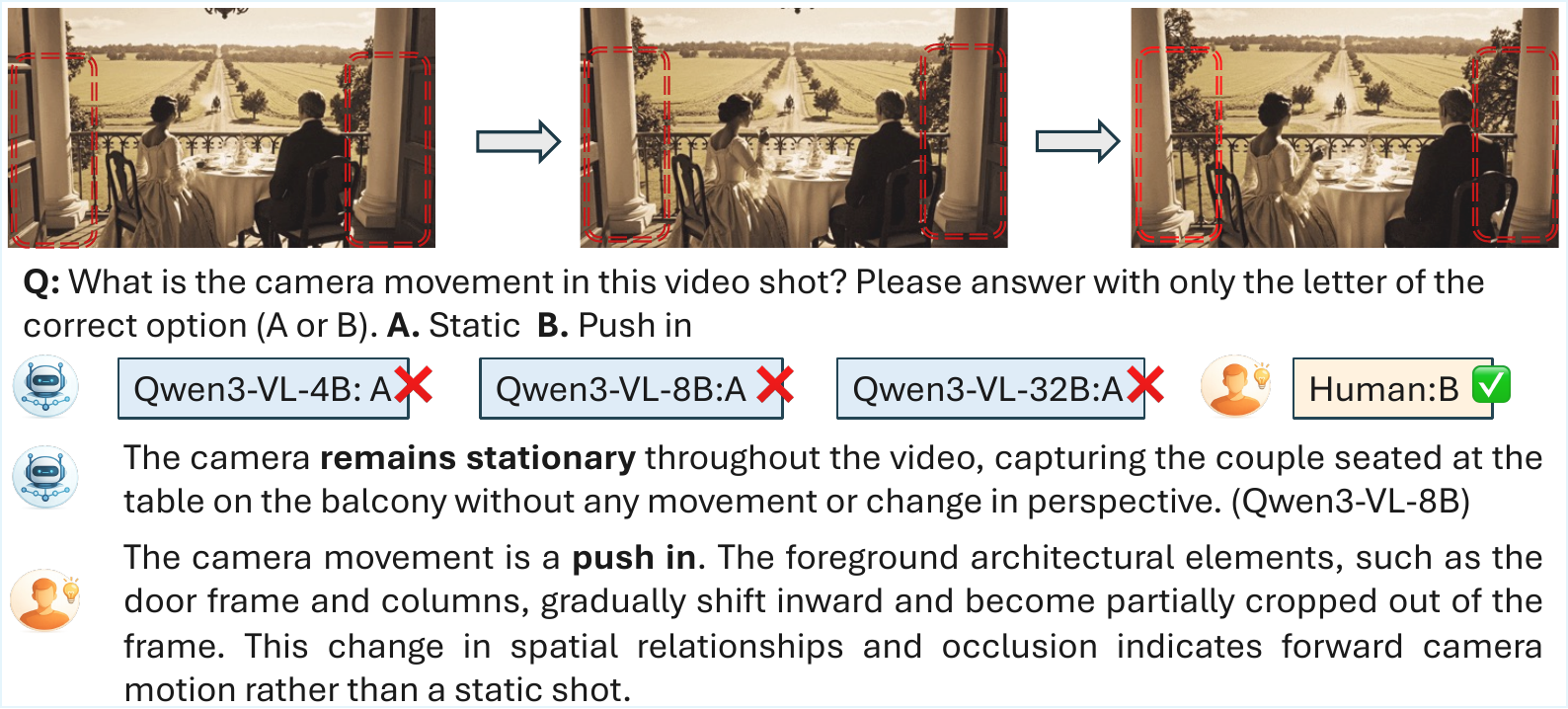}
\caption{An example illustrating  VLMs' limited sensitivity to {small} camera movement.}
\label{fig:fine_grained_cues}
\end{figure}


\subsection{Difficulty in Perceiving {Small} Camera Movement}
We observe that VLMs frequently misclassify dynamic camera movement as `static', particularly when the movement intensity is moderate, suggesting limited sensitivity to inter-frame differences. In contrast, such differences are readily perceptible to humans. As shown in Fig.~\ref{fig:fine_grained_cues}, humans can easily infer that the depicted camera movement is `push in' by perceiving and localizing changes in the foreground, whereas Qwen3-VL models~\cite{bai2025qwen3} consistently predict these videos as `static'. To further quantify this issue, we focus on the `push in' category and manually divide the corresponding videos into three levels of camera movement: low, medium, and high. We then construct a binary-choice question-answering task to evaluate whether models can detect the camera movement.
As shown in Tab.~\ref{tab:pushin_frames_intensity}, Qwen3-VL-32B and Gemini-3-Pro perform near or below random chance under low intensity motions, revealing a substantial gap between VLMs and human perception.

\begin{table}[tb]
\centering
\caption{VLM accuracy on `push in' \textit{vs.}\ three levels of camera movement intensity and the number of frames fed to VLMs.}
\label{tab:pushin_frames_intensity}
\scriptsize
\setlength{\tabcolsep}{2.8pt}
\begin{tabular}{lccc|ccc|ccc}
\toprule
 & \multicolumn{3}{c}{Frames=8/Fps=1}
 & \multicolumn{3}{c}{Frames=16/Fps=2}
 & \multicolumn{3}{c}{Frames=32/Fps=4} \\
\cmidrule(lr){2-4} \cmidrule(lr){5-7} \cmidrule(lr){8-10}
Model 
 & Low & Medium &High
 & Low &  Medium &High
 & Low &Medium& High \\
\midrule
Random&50.00&50.00 &50.00 &50.00&50.00&50.00&50.00 &50.00\ &50.00\\
\midrule
Qwen3-VL-32B~\cite{bai2025qwen3} &41.67  & 81.90 & 100.00 & \textbf{52.08 }& 86.67 & 100.00&45.83  &\textbf{88.57}&\textbf{100.00 }\\
Gemini-3-Pro~\cite{gemini} &29.17&47.62&85.37&\textbf{31.25}&56.19&\textbf{100.00}&29.17&\textbf{58.10}&92.68\\
\bottomrule
\end{tabular}
\end{table}

\subsection{Confusion Between Translation and Rotation}

\begin{figure}[t]
  \centering
\includegraphics[width=0.99\linewidth]{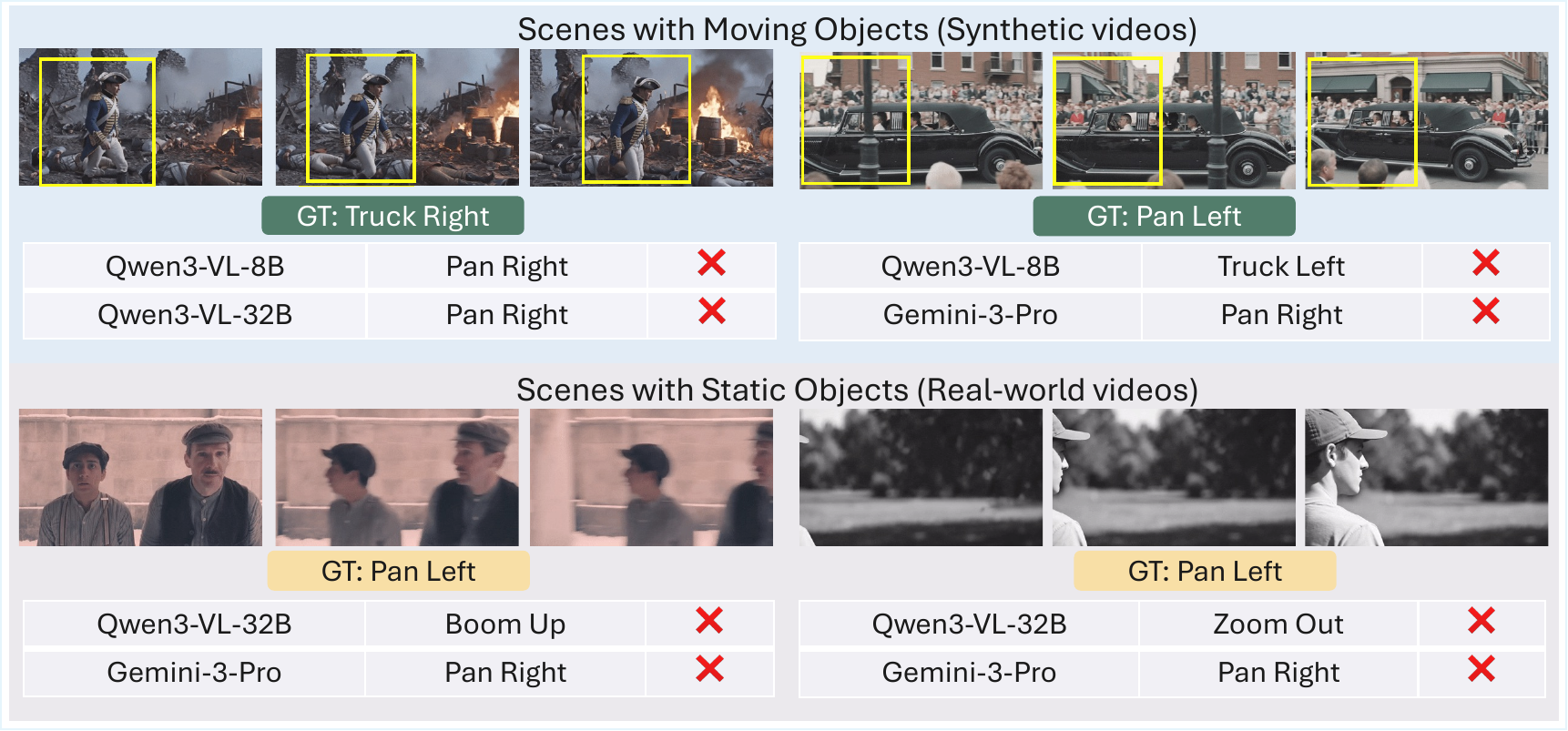}
  \caption{Examples of translation-rotation and left-right confusion.
  }
  \label{fig:example_2}
\end{figure}

\begin{figure}[tb]
  \centering
\includegraphics[width=0.99\linewidth]{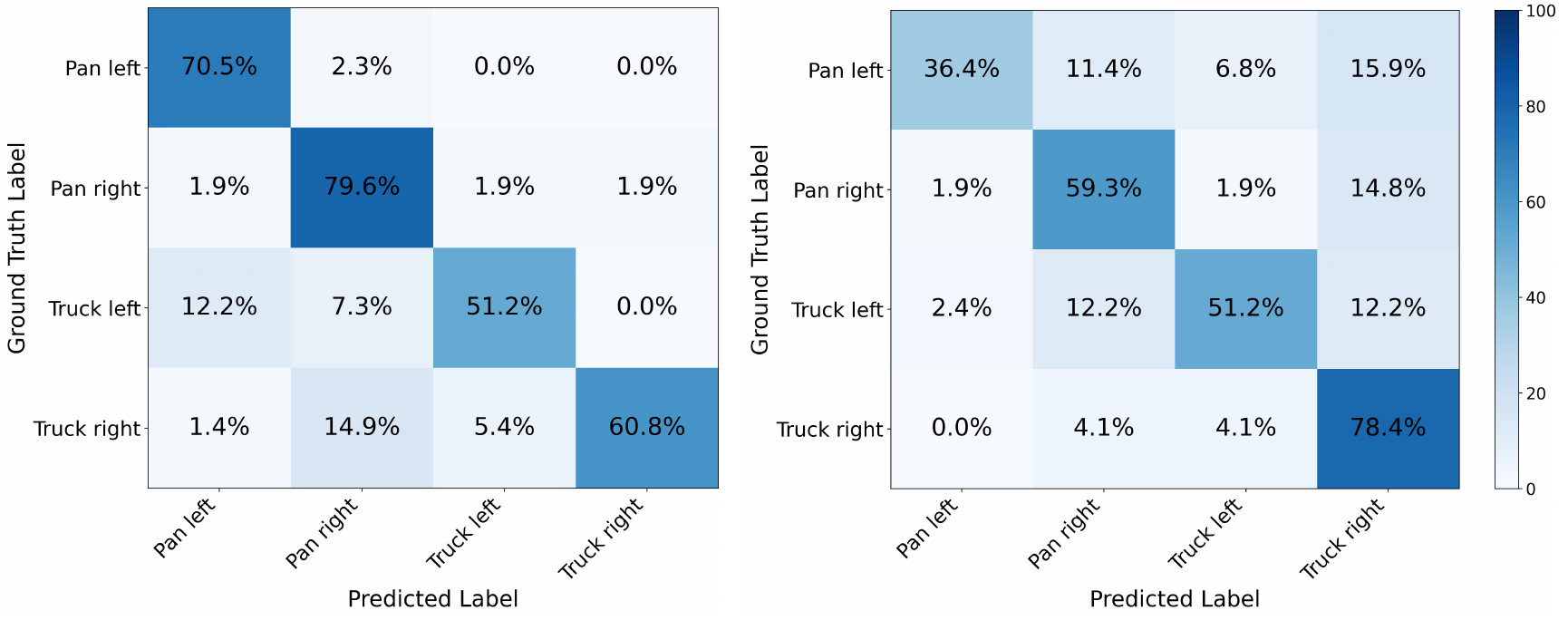}
  \caption{Partial confusion matrices for Qwen3-VL-32B (left) and Gemini-3-Pro (right), illustrating translation-rotation and left-right confusion. (Full matrix in supplement)}
  \label{fig:confusion_metrix}
\end{figure}

\begin{figure}[t]
  \centering
\includegraphics[width=0.99\linewidth]{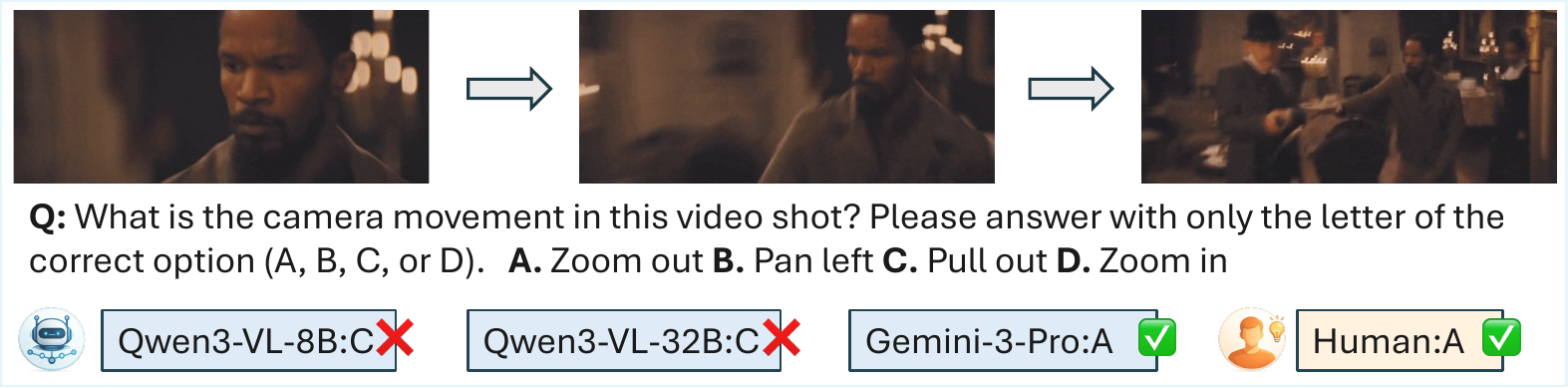}
\caption{A failure case in distinguishing optical `{zoom out}' from physical `{dolly out}'.}
  \label{fig:zoom}
\end{figure}

\begin{figure}[tb]
  \centering
\includegraphics[width=0.99\linewidth]{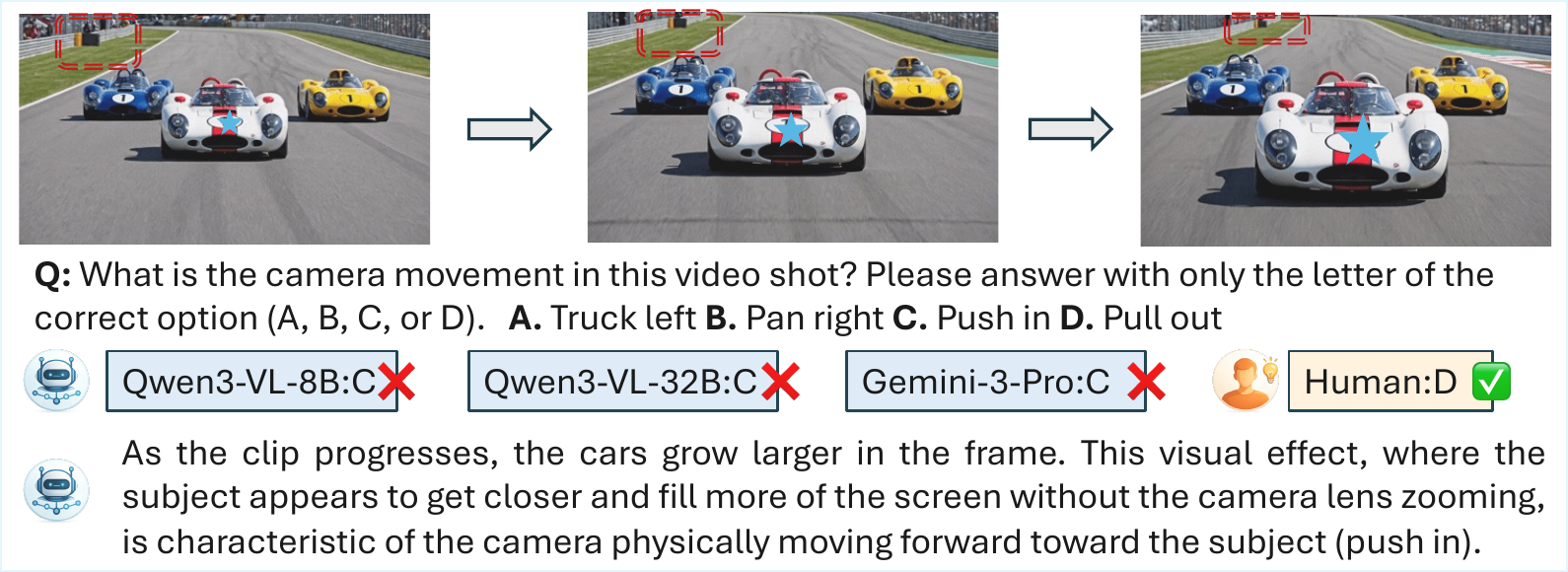}
  \caption{Object motion challenges current VLMs' inference about camera movement. 
  }
  \label{fig:example_3}
\end{figure}

Beyond fine-grained sensitivity, VLMs also frequently struggle to distinguish between camera translation and rotation.  As illustrated in Fig.~\ref{fig:example_2}, when presented with a tracking shot of a walking man (top left), which is a purely lateral camera translation (`{truck right}'), both Qwen3-VL-8B and 32B models misclassify the movement as a horizontal rotation (`{pan right}'). Conversely, in the top right, where the camera rotates horizontally to follow a moving vintage car (`{pan left}'), Qwen3-VL-8B exhibits the exact opposite error, mispredicting the rotation as a translation (`{truck left}'). These reciprocal failures strongly indicate that current VLMs predominantly rely on superficial 2D shifts rather than developing a genuine understanding of 3D camera mechanics and spatial geometry.

\subsection{Left-Right Confusion}

VLMs also often mistake camera's moving left as moving right, and vice versa. A representative example is in the bottom row of Fig.~\ref{fig:example_2}. Although the camera pans left to reveal more details on the left side, Gemini-3-Pro predicts the opposite direction (pan right). We further validate this observation using the confusion matrices in Fig.~\ref{fig:confusion_metrix}. Off-diagonal mass appears between opposite directions within the same movement, particularly for Gemini-3-Pro.

\subsection{Confusion Between Optical Zoom and Physical Dolly}
A prominent blind spot for current VLMs is in their limited ability to distinguish optical scaling from physical movement. Although both produce a centered expansion or contraction, their underlying mechanics differ. Camera translation introduces depth-dependent parallax that shifts the relative positions of foreground and background elements, whereas focal-length changes uniformly magnify the scene while preserving perspective. As illustrated in Fig.~\ref{fig:zoom}, models fail to exploit these cues, often misclassifying optical zoom as physical translation. This suggests that VLMs rely on superficial 2D scaling cues that are insufficient in differentiating camera intrinsics (focal changes) from extrinsic movement.

\subsection{Confusing Object Movement with Camera Movement}
We identify another critical failure mode: VLMs rely on salient foreground dynamics rather than global background cues to infer camera movement. When objects move independently of the camera, they induce local 2D size or position changes, which VLMs often misinterpret as global camera movement, illustrated in Fig.~\ref{fig:example_3}. Although the camera translates backward, the racing cars move forward faster than the camera, causing them to occupy more of the frame over time.  VLMs incorrectly interpret this `subject growing larger' cue as `{push in}', indicating that they behave more like localized object trackers than global spatial observers when tasked with natural language camera movement understanding.

\section{ACaM: A Benchmark for Natural Language Understanding of \textit{A}tomic \textit{Ca}mera \textit{M}ovement}

Findings in \cref{sec:why_hard} indicate that camera movement understanding is a nontrivial task that deserves careful and systematic study. Thus, in order to promote and facilitate future research on it, we compile an extensive benchmark (dubbed ACaM for Atomic Camera Movement) following a taxonomy derived from cinematography, comprising both real-world and synthetic videos. 

\begin{table}[t]
  \caption{Comparison with camera motion understanding benchmarks. For benchmarks not designed for this task, statistics are reported only for clips involving camera motion.}
  \label{tab:comparison}
  \centering
  \scriptsize
    \setlength{\tabcolsep}{3pt} 
  \begin{tabular}{@{}lcccccccc@{}}
    \toprule
    \multirow{2}{*}{Benchmark}
    & \multicolumn{2}{c}{\#Domain}
    & \multicolumn{2}{c}{\#Videos}
    & \multicolumn{2}{c}{\#QA Format}
   & {\#Human}  & \#Labels \\
    \cmidrule(lr){2-3}
    \cmidrule(lr){4-5}
    \cmidrule(lr){6-7}
     \cmidrule(lr){8-9}
      \cmidrule(lr){9-9}
    & Real & Syn
    & Train & Test
    & Yes/No & MCQ
     & Performance
     & Number
     \\
    \midrule
    Cinematic2K~\cite{li2024can}&\cmark  & \xmark  & \xmark & 1,047 & \xmark & \xmark   &\xmark &11 \\
    CameraBench~\cite{camerabench} &\cmark  & \xmark  & 1,402 & 1,071 & \cmark  & \xmark   &\cmark  & 50 \\
    ShotBench-Subset~\cite{shotbench} &  \cmark & \xmark  &1,058  & 464 &  \xmark & \cmark   &\xmark &17  \\
CineTechBench-Subset~\cite{cinetechbench}     &\cmark  & \xmark  & \xmark & 120 & \xmark &  \cmark &\xmark &15 \\
   MotionBench-Subset~\cite{motionbench}       &\cmark  &\xmark   &5,000  &  385&\xmark  &  \cmark  &\xmark  &- \\
      FavorBench-Subset~\cite{favorbench}      & \cmark & \xmark  & 17,000 &546  &\xmark  & \cmark  &\xmark &- \\
       \midrule
      ACaM (Ours)       &  \cmark& \cmark &  24,321 & {2,602} & \cmark    & \cmark & \cmark &17 \\
    \bottomrule
  \end{tabular}
\end{table}

\subsection{Camera Movement Taxonomy}
We categorize camera movement into 17 distinct classes (see \cref{tab:motion_taxonomy}) following the cinematography literature~\cite{nielsen2007camera,keating2019dynamic}. 
{CameraBench ~\cite{camerabench} introduces a comprehensive taxonomy of over 50 motion primitives, while our evaluation focuses on 17 atomic, intentional camera movements that can be unambiguously described in natural language.}
Furthermore, our benchmark provides a more comprehensive set of categories than Cinematic2K~\cite{li2024can} and CineTechBench~\cite{cinetechbench}, spanning five core dimensions: physical translations, angular rotations, focal length changes, static shots, and object-centric movements. \cref{tab:comparison} compares our benchmark with existing camera movement and motion understanding benchmarks. {Notably, our ACaM is the first to include synthetic videos and human performance as a reference baseline for direct human–model comparison.}

\begin{figure}[t]
  \centering
   \includegraphics[width=0.99\linewidth]{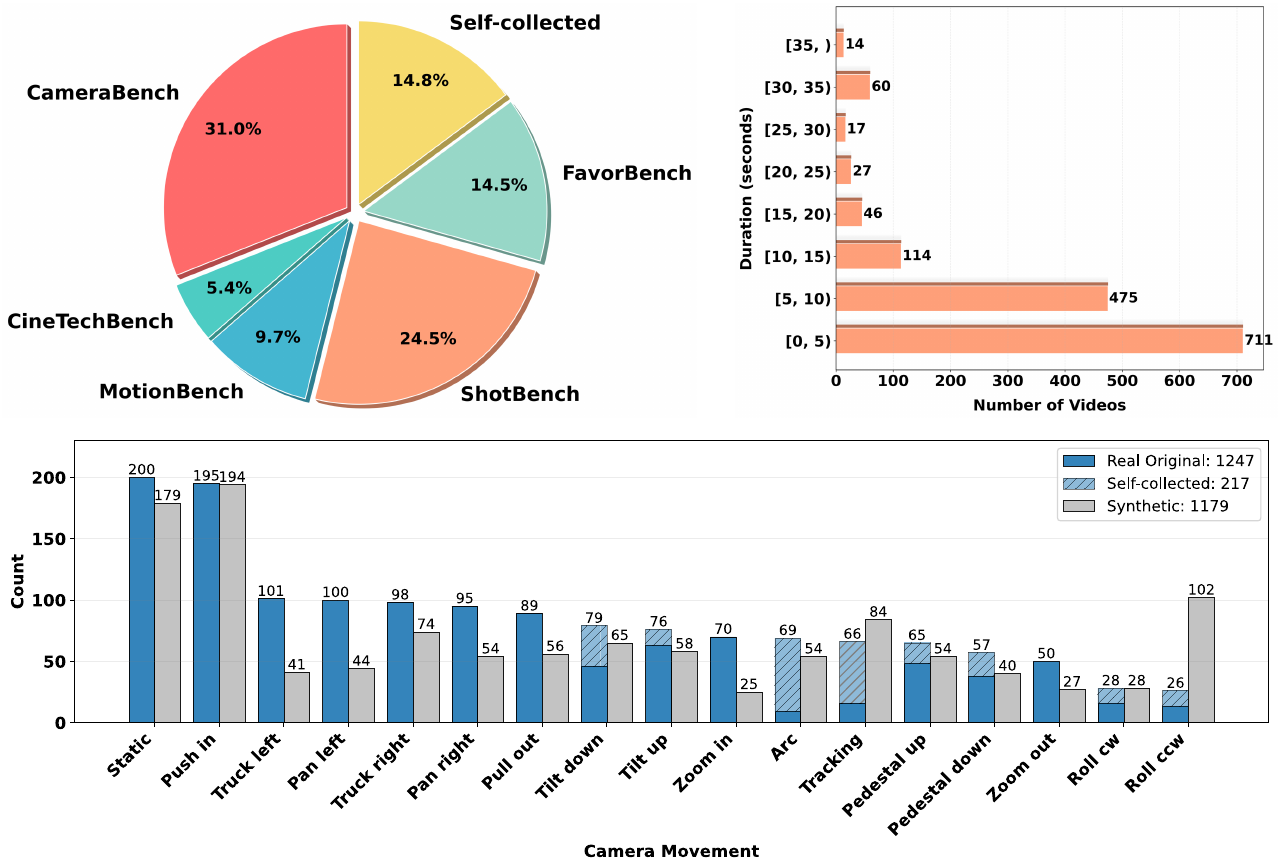}
\caption{Dataset statistics of our ACaM evaluation benchmark. 
Top: Source composition and duration distribution of real-world videos. Bottom: Class-wise camera movement distribution of real-world and synthetic videos.}
  \label{fig:statistic}
\end{figure} 

\subsection{Real-World Videos} 
\textbf{Data construction pipeline.}
To construct a diverse and high-quality set of real-world videos, we unify several existing motion understanding benchmarks~\cite{camerabench,shotbench,cinetechbench,motionbench,favorbench}. 
Given the heterogeneous annotation formats and motion categories across these datasets, we apply tailored filtering and refinement strategies to each source. We first use the binary question-answering subset from CameraBench~\cite{camerabench}, which is designed for camera motion understanding. To focus on atomic movement recognition, we retain only videos with a single affirmative label. 
We then use Gemini-3-Pro~\cite{gemini} to parse movement descriptions from captions and remove detected secondary movements from the negative answer options, ensuring the QA task remains centered on the primary camera movement.
For the cinematography benchmarks of ShotBench~\cite{shotbench} and CineTechBench~\cite{cinetechbench}, we extract subsets related to camera movement and retain only evaluation questions involving a single camera movement. 
Similarly, for fine-grained motion understanding benchmarks (MotionBench~\cite{motionbench} and FavorBench~\cite{favorbench}), we extract the camera-related QA subsets and remove questions involving object motion. 
These datasets present an interesting challenge, among others, since they often require identifying camera movement for a specific event or timestamp (see examples in the supplementary material). Finally, to enrich underrepresented camera movement classes (see \cref{fig:statistic}, bottom), we search for and curate YouTube videos. 
The assembled dataset is manually verified by graduate students with cinematography training to remove incorrectly labeled or ambiguous samples across all sources (see supplementary materials for details).

\noindent\textbf{Dataset Statistics.} \cref{fig:statistic} summarizes some statistics of our real-world video set. After filtering and refinement, the final benchmark contains {1,423} videos and {1,464} QA pairs, covering 17 camera movement categories aggregated from six different sources. More than 80\% of the videos are shorter than 10 seconds, ensuring that the evaluation primarily focuses on the dominant, atomic camera movement behind each video. Regarding the class distribution, `{static}' and `{push in}' account for a slightly larger portion than the others, while the `{rolling}' classes are underrepresented even after we enrich them with YouTube videos. 

\subsection{Synthetic Videos }
\begin{figure}[t]
  \centering
   \includegraphics[width=0.99\linewidth]{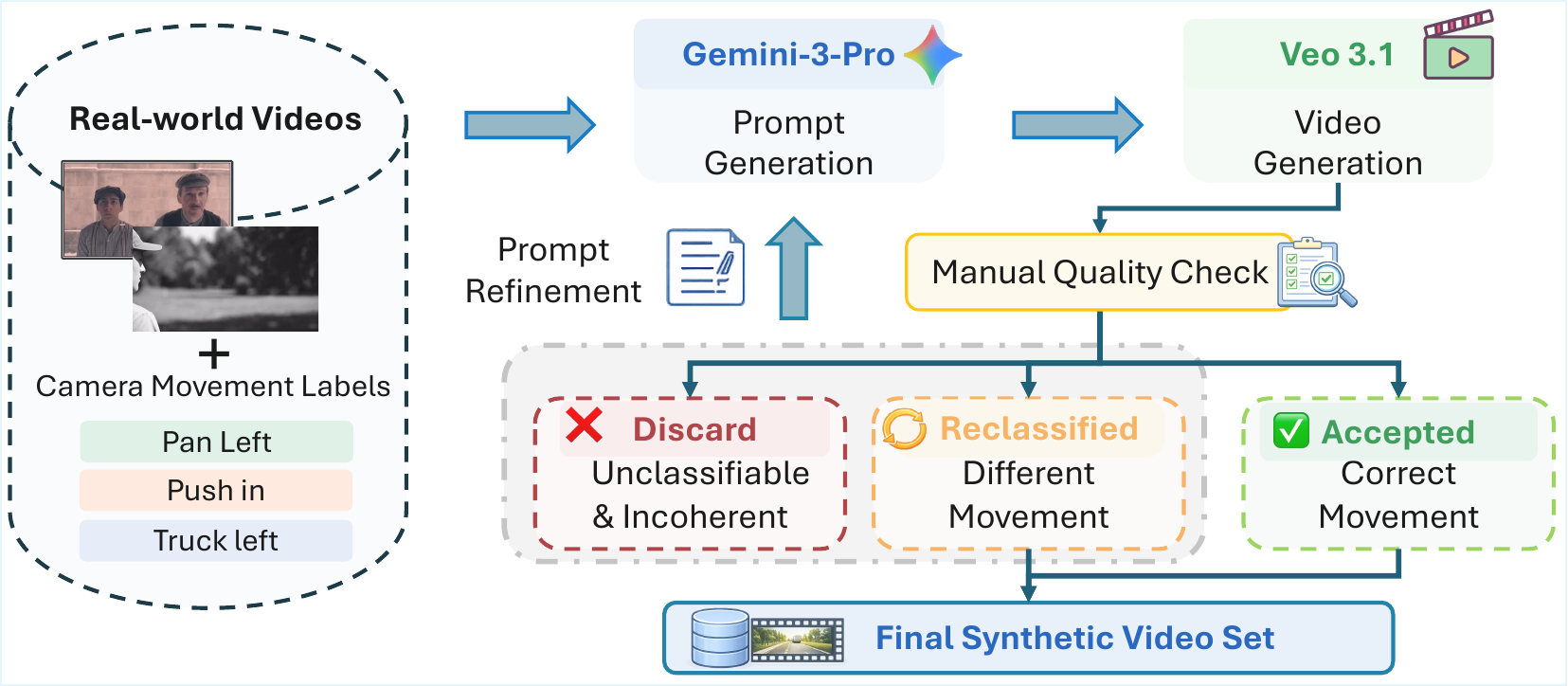}
  \caption{Pipeline for generating synthetic videos of various camera movement classes.
  }
  \label{fig:syn_pipe}
\end{figure}

In addition to real-world videos, ACaM also incorporates AI-generated videos, mimicking the automatic rating of generative models using VLMs. \cref{fig:syn_pipe} shows the pipeline for generating synthetic videos using Veo 3.1~\cite{veo}---our lab currently has research credits for no platform but Google Cloud.

To construct the synthetic video set, we start from our realistic videos for each camera movement class.
For each clip, we feed the raw video and its camera movement label to Gemini-3-Pro~\cite{gemini} while instructing it to analyze the video and produce a structured video generation prompt. The resulting prompts are then fed to Veo 3.1, which performs video generation to synthesize clips aligned with the camera movement in the prompt. After generation, we conduct a manual quality-control pass as follows. Each clip is assigned to one of three outcomes: \textit{accepted} if it exhibits the intended camera movement, \textit{reclassified} if it depicts a different camera movement label in our taxonomy, or \textit{discarded} if the movement is unclassifiable or temporally incoherent (e.g., abrupt transitions). For discarded or reclassified cases, we let Gemini-3-Pro rewrite the video generation prompt to reduce ambiguity and enhance clarity of the intended camera movement. We then send the refined prompts to Veo 3.1 for a second round of video generation.

{In general, `zoom', `arc', and `roll (clockwise)' are more difficult to generate than the other camera movement classes.} For these categories, we employ a separate prompt generation script that conditions solely on the designated camera movement label, without any anchor on real-world videos. The supplementary material details the whole pipeline with examples.


\noindent\textbf{Data Statistics.}
The synthetic videos are four seconds long. Their class-wise distribution is shown on the bottom of Fig.~\ref{fig:statistic}.
Despite the iterative pipeline, certain `zoom' and `roll' classes of videos remain difficult to generate. Interestingly, Veo 3.1 has a systematic bias toward `{roll counterclockwise}'; many instructions for `{roll clockwise}' actually lead to videos of `{roll counterclockwise}'.
`Zoom'-related instructions often result in `static' shots or `dolly' movement. 


\section{Towards VLMs that understand camera movement in natural language}
To enhance VLM capabilities in camera movement understanding, we construct an instruction-tuning set consisting of 27K samples by drawing from 45K raw video clips, enhanced with targeted camera movement augmentation. 



\begin{table}[tb]
\centering
\caption{Sources and statistics of our instruction-tuning set (MCQ: Multi-Choice QA)}
\label{tab:data_sources}
\scriptsize
    \setlength{\tabcolsep}{3pt}
\begin{tabular}{lcccc}
\toprule
Dataset &\#Extracted Clips & \#Annotation & \#Label Source & \#Label Num. \\
\midrule
CameraBench~\cite{camerabench} & 723& Motion Caption & Human Annotation & 16\\
ShotBench~\cite{shotbench} & 375 & MCQ Format & Human Annotation & 14 \\
SpatialVID~\cite{spatialvid} & 29,733 &  Movement Labels & Estimated Poses & 12\\
MultiCamVideo~\cite{MultiCamVideoDataset} & 12,048 & Camera Pose& Synthetic Trajectory &13  \\
GenDoP~\cite{gendop} & 2,438 &Movement Caption & Estimated Poses&6 \\
\midrule
\textbf{Overall} &45,317 & MCQ Format & Mixed &17\\
\bottomrule
\end{tabular}
\end{table}

\subsection{Data Processing and Filtering}

While CameraBench~\cite{camerabench} and ShotBench~\cite{shotbench} provide videos with camera movement labels, their training sets are small (approximately 1K video clips; see \cref{tab:comparison}). We pool them and further incorporate video clips from three datasets that contain synthetic or estimated camera trajectories. Specifically, we use a subset of SpatialVID~\cite{spatialvid}, which provides estimated camera poses, despite being highly dominated by the `push in' movement. MultiCamVideo~\cite{MultiCamVideoDataset} is a dataset of synthetic videos rendered in Unreal Engine~5 with controlled camera trajectories, and GenDoP~\cite{gendop}, which provides natural language descriptions of camera movements. \cref{tab:data_sources} summarizes the data sources and labels.

\begin{figure}[tb]
  \centering
  \includegraphics[width=0.99\linewidth]{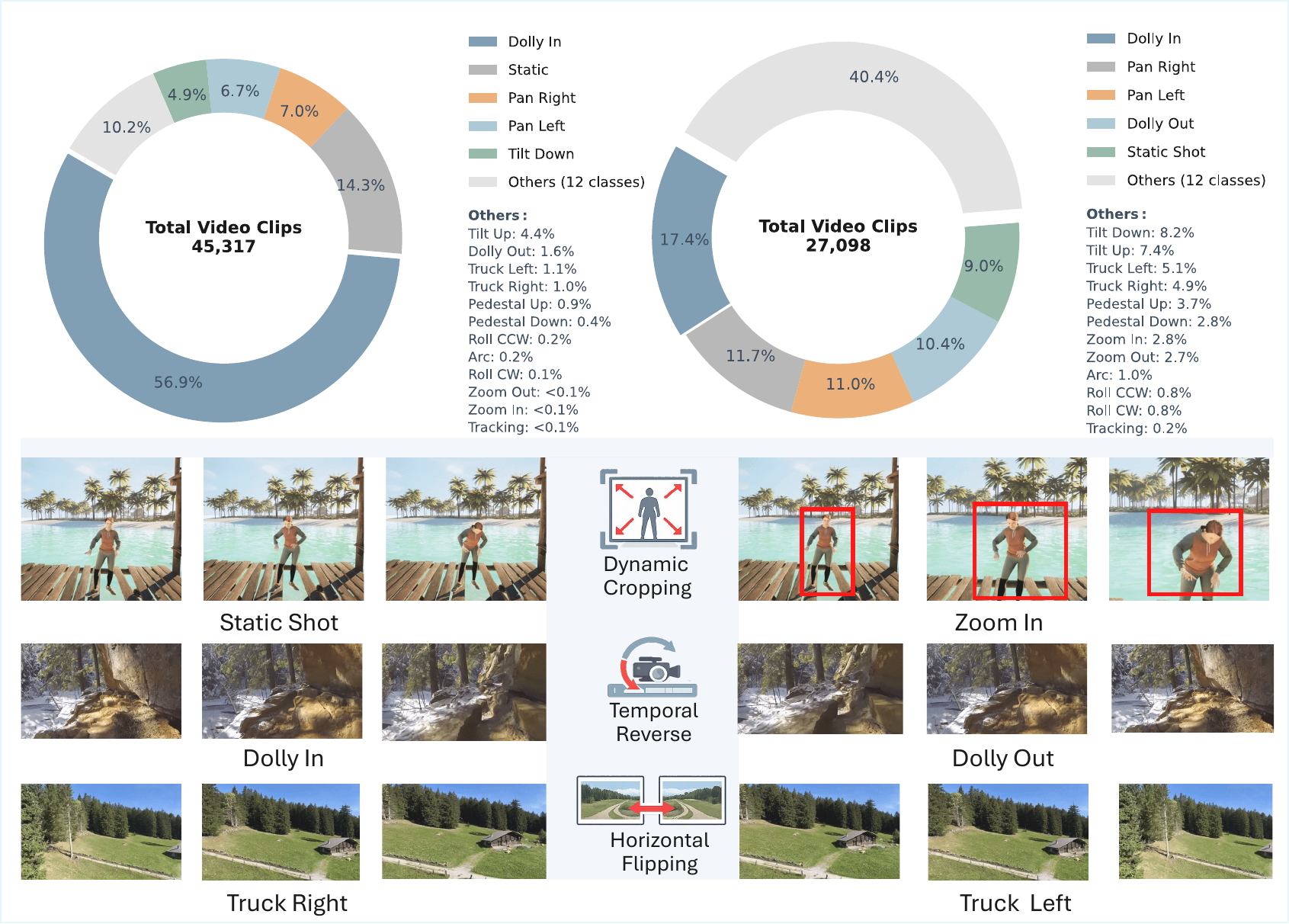}
  \caption{Targeted data augmentation for distribution balancing. Top: The shift from an unbalanced 45K raw dataset to a refined 27K instruction-tuning set. Bottom: The augmentation operators used to synthesize samples for tail classes.
  }
  \label{fig:training_samples}
\end{figure}

Our training corpus prioritizes atomic camera movement, which we believe is the first step towards complete natural language understanding for various, potentially compound, and long-form camera movements. From ShotQA~\cite{shotbench}, we retain only the subset of single movements. For CameraBench~\cite{camerabench} and GenDoP~\cite{gendop}, we employ GPT-5-nano~\cite{gpt5} to parse camera movement descriptions from their video captions and convert them to multi-choice question-answering (MCQ) tasks. For MultiCamVideo~\cite{MultiCamVideoDataset}, frame-wise camera poses are projected into semantic movement labels using geometric heuristics, retaining only the segments containing a single camera movement. In SpatialVID~\cite{spatialvid}, due to noisy VLM-generated global captions, we instead adopt their frame-wise camera movement descriptions as the groundtruth. Since pose annotations in SpatialVID and GenDoP are derived from geometry-based estimation and prone to artifacts, we further perform manual verification to correct any labeling errors. After filtering, the final training corpus contains 45,317 video clips in the MCQ format, as summarized in Tab.~\ref{tab:data_sources}. More details are provided in the supplementary material.

\begin{table}[t]
   \caption{Results of various models on our ACaM's real-world videos}
  \label{tab:real-world videos}
  \centering
  \scriptsize
   \setlength{\tabcolsep}{4.5pt} 
  \begin{tabular}{l|cccccc|cc}
    \toprule
    Model  & Static& Rot. &Trans.& Zoom &Arc &Track & Overall & Avg \\
    \midrule
  Random Guess&24.50&24.28&24.00&24.54&24.78&25.00&24.27&24.37\\
  Human Performance&99.00&93.07&90.39&90.83&97.10&96.97&93.44 &93.14\\
     \midrule
    \multicolumn{9}{c}{\textbf{Visual Geometry Models}} \\
    \midrule
    Mega-SaM~\cite{megasam}&89.55 & \textbf{74.94}& 54.16&-- &--&--&--&-- \\
    ViPE~\cite{vipe}& 61.81& 49.71&\textbf{73.17} &-- &--&--&--&-- \\
  \midrule
    \multicolumn{9}{c}{\textbf{Spatial VLMs}} \\
      \midrule
          G$^2$VLM-2B~\cite{hu2025g2vlm}&9.50&4.52&4.52&7.44&5.71&7.58&5.66&5.24 \\
            Spatial-MLLM~\cite{wu2025spatial}&59.00& 16.98&31.72&41.67&24.29&77.27&34.22&29.68\\
    VLM-3R-7B~\cite{fan2025vlm}&\textbf{94.50}&20.69&28.55&35.83&51.43&89.39&40.22&35.02 \\
      \midrule
     \multicolumn{9}{c}{\textbf{General VLMs}} \\
       \midrule
     Qwen3-VL-4B ~\cite{bai2025qwen3}   & \textit{93.50}   &53.47&34.21&61.67&56.52&80.30&53.01&46.96  \\
       Qwen2.5-VL-7B~\cite{bai2025qwen2} & 81.00 &  33.91&34.71&58.33&66.67&92.42&46.86 &45.62\\
    Qwen3-VL-8B  ~\cite{bai2025qwen3}  &85.00  &65.84  &39.50 &61.67&79.71&74.24&58.27&55.25\\
    LLaVa-OV-7B~\cite{llavaov}   & 92.50 &  28.71&25.79&24.17&62.32&75.76&39.55&34.85  \\
    InternVideo2.5-8B~\cite{internvideo2.5}&82.50&31.44&28.43&64.17&50.72&92.42&43.51 &41.07\\
    InternVL3.5-8B~\cite{internvl3.5} &89.00&48.51&30.58&58.33&40.58&86.36&48.77 &46.15\\
    InternVL3.5-14B~\cite{internvl3.5} &81.00&43.56&39.83&63.33&55.07&89.39&51.37 &47.21\\
    Qwen3-VL-32B  ~\cite{bai2025qwen3}  &92.00 &64.11 & 47.93&64.17&57.97&86.36&61.95&58.66\\
       {GPT-5}~\cite{gpt5}  & 61.50&66.58&54.55&61.67&68.12&80.30&61.20& 61.54\\
   Gemini-3-Pro~\cite{gemini}  & 83.50 & 64.36 &60.84 &65.00&82.61&96.97&68.24&66.83\\
    {Gemini-3.1-Pro}~\cite{gemini}  & 83.92 & 63.34 &64.00 &62.39&76.81&95.38&68.44&67.73\\
                \midrule
    \multicolumn{9}{c}{\textbf{Camera Movement Specialized VLMs}} \\
      \midrule
          CameraModel-7B~\cite{camerabench}    & 55.50 &68.81 &47.93& 45.83&\textbf{91.30}&\textbf{98.48}&58.88& 60.56 \\
    ShotVL-7B~\cite{shotbench}     & 91.00 &60.40 &51.07 &32.50&68.12&98.48&60.52 &57.09\\
   CamReasoner-7B~\cite{camreasoner}    &76.50 & 41.58&32.89&59.17&39.13&80.30&45.83&44.99 \\
      \midrule
          \rowcolor{blue!5}
   Our SFT Qwen3-VL-4B & 76.00&\textit{71.53}&\textit{64.13}&\textit{76.67}&78.26&93.94&\textit{70.83}&\textit{72.27 }\\
      \rowcolor{blue!5}
   Our SFT Qwen3-VL-8B & 85.50&70.05&64.13&\textbf{84.17}&\textit{84.06}&\textit{96.97}&\textbf{72.75}&\textbf{74.28} \\
    \bottomrule
  \end{tabular}
\end{table}
\subsection{Data Sampling and Augmentation}
The top-left panel of Fig.~\ref{fig:training_samples} shows that the integrated instruction-tuning set exhibits a significant long-tail distribution, which could introduce strong bias to VLMs if we fine-tune models directly on this set. Hence, we re-sample the videos and design targeted data augmentation, striving to balance the videos across camera movement categories. Specifically, we address the scarcity of `zoom' by progressively cropping static frames and resizing them to the original resolution, yielding approximately 1K augmented videos. We employ temporal reversal to transform a subset of `dolly in' sequences into their `dolly out' counterparts, yielding 2K augmented videos. Similarly, horizontal flipping is applied to `truck' movement, producing about 1.7K additional videos. Finally, to enrich the extremely scarce `roll' classes, we synthesize rolling dynamics by applying continuous affine rotations to static video shots. By combining the re-sampling and video augmentation strategies, we arrive at a refined, more balanced training corpus of 27K high-quality video-MCQ pairs.

\begin{table}[tb]
  \caption{Results of various models on our ACaM's synthetic videos}
  \label{tab: generated videos}
    \centering
  \scriptsize
     \setlength{\tabcolsep}{4.5pt} 
  \begin{tabular}{l|cccccc|cc}
    \toprule
    Model  &  Static& Rot.&Trans. & Zoom &Arc &Track & Overall & Avg \\
    \midrule
  Random Guess&25.00&25.00&25.00&25.00&25.00&25.00&25.00&25.00 \\
  Human Performance& 98.88& 97.72&98.04 & 96.15& 100.00& 97.62& 98.05&97.61  \\
     \midrule
    \multicolumn{9}{c}{\textbf{Visual Geometry Models}} \\
    \midrule
    Mega-SaM~\cite{megasam}& 95.53& \textbf{76.35}& 61.66&-- &--&--&--&-- \\
    ViPE~\cite{vipe}&68.16 & 52.42& \textbf{78.65}&-- &--&--&--&-- \\
  \midrule
    \multicolumn{9}{c}{\textbf{Spatial VLMs}} \\
        \midrule
            G$^2$VLM-2B~\cite{hu2025g2vlm} &4.44&2.91&6.99&1.92&5.45&13.10&5.54&4.51\\
            Spatial-MLLM~\cite{wu2025spatial}&68.33&29.65&27.07&50.00&43.64&94.05&40.75&36.04 \\
    VLM-3R-7B~\cite{fan2025vlm}&\textbf{99.44}&18.90&43.45&51.92&47.27&89.29 &48.68&38.94\\     \midrule
    \multicolumn{9}{c}{\textbf{General VLMs}} \\
         \midrule
    Qwen3-VL-4B ~\cite{bai2025qwen3}   & \textit{97.77}& 54.42&42.92&73.08&40.74&72.62&58.02& 53.68 \\
       Qwen2.5-VL-7B~\cite{bai2025qwen2} & 91.06 & 52.99 &50.54 &82.69&61.11&94.05& 62.43&58.50\\
    Qwen3-VL-8B  ~\cite{bai2025qwen3}  & 90.50 &62.39  &45.10 &\textit{80.77}&55.56&83.33&61.92&60.91\\
    LLaVa-OV-7B~\cite{llavaov}   & 96.09&39.60&37.25&36.54&51.85&86.90&51.06&41.59\\
    InternVideo2.5-8B~\cite{internvideo2.5}&94.97&44.73&31.15&75.00&27.78&\textit{91.67}&50.98 &46.71\\
    InternVL3.5-8B~\cite{internvl3.5} &96.09&47.29&33.99&67.31&35.19&73.81&51.74 &47.73\\
    InternVL3.5-14B~\cite{internvl3.5} &88.27&41.60&47.93&76.92&31.48&86.90&55.47 &51.15\\
    Qwen3-VL-32B  ~\cite{bai2025qwen3}  & 91.62 &  68.95& 55.77&82.69&51.85&67.86&67.01&64.17\\
       {GPT-5}~\cite{gpt5}  & 64.80&69.80 &57.95&82.69&53.70&63.10&63.78& 65.09\\
    Gemini-3-Pro~\cite{gemini}  & 89.94&61.54 & 68.63&73.08&53.70&88.10&70.65& 66.27\\
     {Gemini-3.1-Pro}~\cite{gemini}  & 92.07 & 62.00&74.07&69.23&55.56&85.37&73.01&69.58\\
   %
   \midrule
    \multicolumn{9}{c}{\textbf{Camera Movement Specialized VLMs}} \\
      \midrule
    CameraModel-7B~\cite{camerabench}    & 58.10 & 62.68&51.63& 71.15&\textbf{88.89}&98.81&61.83& 68.45 \\
        ShotVL-7B~\cite{shotbench}     & 96.09 &69.80  & 63.62&63.46&29.63&\textbf{98.81}& 71.33&64.49\\
    CamReasoner-7B~\cite{camreasoner}    &84.92 & 51.28&42.70&67.31&55.56&77.38&55.81&53.74\\
      \midrule
          \rowcolor{blue!5}
   Our SFT Qwen3-VL-4B & 82.68&\textit{74.36}&66.67&78.85&68.52&84.52&73.28&74.40\\
         \rowcolor{blue!5}
    Our SFT Qwen3-VL-8B & 94.41&72.36&\textit{74.51}&\textbf{82.69}&\textit{68.52}&91.67&\textbf{78.20}& \textbf{77.44}\\
    \bottomrule
  \end{tabular}
\end{table}
\section{Experiments}
We apply supervised fine-tuning (SFT) to Qwen3-VL-4B and 8B over our training set. We also test geometry-based methods~\cite{megasam,vipe}, spatial VLMs~\cite{wu2025spatial,hu2025g2vlm, fan2025vlm}, general VLMs~\cite{bai2025qwen2,gemini,gpt4o,llavaov,internvl3.5,internvideo2.5,bai2025qwen3}, and prior camera movement specialized VLMs~\cite{shotbench,camerabench,camreasoner}. The supplementary material reports the detailed training and evaluation setups, and Tab.~\ref{tab:real-world videos} and Tab.~\ref{tab: generated videos} show the quantitative results on our benchmark's real and synthetic videos, respectively.  We report both the overall accuracy and the average accuracy across the 17 camera-movement classes (additional results for the binary QA setting are provided in the supplementary material). The main observations are as follows.

    
Geometry-based models~\cite{megasam,vipe} demonstrate complementary strengths to VLMs; Mega-SaM excels in static scenes and camera rotations, whereas ViPE performs the best on translations. However, neither achieves robust performance across all classes, revealing the limitations of geometry-based methods to natural language understanding of camera movement. Indeed, these methods do not explicitly model semantics and often assume fixed intrinsics when estimating camera pose, limiting their applicability to predict `zoom', `tracking', and `arc'. 
    
SpatialVLMs~\cite{wu2025spatial,hu2025g2vlm,fan2025vlm} incorporate geometry-aware learning objectives when they fine-tune base VLMs. However, their performance on our ACaM is surprisingly low, most likely because they were specialized for the spatial understanding of the video content only, rather than its spatial relationship with camera movement. Among the general VLMs of no more than 8B parameters, Qwen3-VL~\cite{bai2025qwen3} achieves the strongest performance. Furthermore, Qwen3-VL's 32B model significantly outperforms its 4B and 8B models and is nearly on par with Gemini-3-Pro on synthetic videos. {Gemini-3/3.1-Pro} exhibits a stronger capability to perceive translational movement than other VLMs.  


We next compare our SFT Qwen3-VLs to the existing camera movement specialized VLMs. CameraModel-7B~\cite{camerabench} is competitive, even on par with Gemini-3-Pro on synthetic videos, but it sacrificed its discrimination ability for static shots. In contrast, ShotVL-7B~\cite{shotbench} performs better on translational movement and static shots than other models. Overall, the existing specialized models perform better on object-centric movements than other VLMs, but they all struggle with `translation' and `zoom'. Our SFT Qwen3-VL-4B significantly outperforms the larger 7B VLMs that were previously specialized for camera movement, with 9\% to 19\% gains, and our 8B model boosts the results further, even improving over {Gemini-3.1-Pro} by 10\% and {11}\% on the real and synthetic videos, respectively. However, even the strongest models fall far behind humans, revealing a noticeable gap between human and VLM performance.

\newlength{\oldintextsep}
\setlength{\oldintextsep}{\intextsep}
\setlength{\intextsep}{0pt}

\begin{wraptable}[11]{r}{0.48\textwidth}
\caption{Ablation of sampling, augmentation and different LoRA rank.}
\label{tab:ablation}
\centering
\scriptsize
\setlength{\tabcolsep}{3.5pt}
\begin{tabular}{cccccc}
\toprule
\multirow{2}{*}{Samp.} &
\multirow{2}{*}{Aug.} 
& \multirow{2}{*}{LoRA} 
& \multirow{2}{*}{Model} 
& \multicolumn{2}{c}{Avg.} \\
\cmidrule(lr){5-6}
& & & &Real & Syn \\
\midrule
$\times$ &$\times$ &r=64  & 4B & 48.42 & 58.56 \\
$\checkmark$ &$\times$ &r=64  & 4B & 63.78 & 72.07 \\
$\checkmark$ &$\checkmark$ & r=64  & 4B & 70.88& 75.03 \\
$\checkmark$ &$\checkmark$ & r=128 & 4B & 71.88 & 75.20 \\
$\checkmark$ &$\checkmark$ & r=256 & 4B & 72.27 & 74.40\\
\rowcolor{blue!5}
$\checkmark$ &$\checkmark$ & r=256  & 8B & \textbf{74.28} & \textbf{77.44} \\
\bottomrule
\end{tabular}
\end{wraptable}

\noindent\textbf{Ablation study.} We further analyze the impact of targeted data sampling, augmentation, and LoRA rank in Tab.~\ref{tab:ablation}. Notably, the combination of targeted sampling and augmentation significantly improves performance, increasing accuracy from $58.56\%$ to $75.03\%$ on synthetic videos, which validates the effectiveness of the data balancing strategy. The LoRA rank has a marginal impact on the 4B model, whereas a higher rank yields better performance for the 8B model (results for the 8B model are provided in the supplementary material).

\setlength{\intextsep}{\oldintextsep}

\section{Related Work}
\textbf{Motion Understanding Benchmarks.}
Motion understanding is a fundamental component of video understanding~\cite{motionsight,motionbench,favorbench,camerabench} and plays an important role in many downstream applications. 
MotionBench~\cite{motionbench} and FavorBench~\cite{favorbench} evaluate fine-grained motion perception, covering both camera and object motion. 
CameraBench~\cite{camerabench} focuses specifically on camera motion, providing around 3K expert-annotated videos and introducing a cinematography-inspired taxonomy for systematic evaluation. 
CamReasoner~\cite{camreasoner} further formulates camera motion understanding as a reasoning task, leveraging chain-of-thought supervision and reinforcement learning to improve performance on CameraBench. 
In addition, ShotBench~\cite{shotbench} (RefineShot~\cite{wu2025refineshot}), CineTechBench~\cite{cinetechbench}, and Cinematic2K~\cite{li2024can} construct expert-annotated QA pairs from film content across cinematography-related~\cite{cinescale2,movienet,rao2020unified} dimensions, including camera motion understanding.

\noindent\textbf{Camera Pose Estimation through Visual Geometry Learning.}
Estimating camera parameters and scene geometry from videos is a long-standing problem in computer vision. 
Classical approaches include SLAM~\cite{megasam,vipe} and Structure-from-Motion (SfM)~\cite{schonberger2016structure,pan2024global}, which explicitly estimate camera trajectories and reconstruct 3D structure from multi-view consistency. More recent methods adopt end-to-end learning frameworks to directly predict 3D geometry or camera poses from video inputs, such as CUT3R~\cite{CUTR3}, and VGGT~\cite{wang2025vggt}. 
 Although these approaches recover geometric structure, they generally lack high-level semantic reasoning. 
Recent efforts attempt to bridge geometry and language by incorporating geometric representations into VLMs~\cite{fan2025vlm,wu2025spatial,hu2025g2vlm}.

\section{Conclusion}
This work establishes natural language camera movement understanding as a standalone research task. We instantiate it by introducing ACaM, an extensive benchmark grounded in a two-level cinematographic taxonomy and covering both real-world and synthetic videos. To deliver the best-performing VLM model, we construct a large-scale training corpus with targeted sampling and augmentation. Our fine-tuned VLM-8B outperforms {Gemini-3.1-Pro} by 10\% on real-world videos and {11}\% on synthetic videos. Despite these gains, the persistent gap to human performance underscores the need for further research.

\section*{Acknowledgements}
This work was supported in part by NSF 2540851 and a Gemini Academic Program Award. 

%
%
\bibliographystyle{splncs04}
\bibliography{main}

@article{shotbench,
  title={ShotBench: Expert-Level Cinematic Understanding in Vision-Language Models},
  author={Liu, Hongbo and He, Jingwen and Jin, Yi and Zheng, Dian and Dong, Yuhao and Zhang, Fan and Huang, Ziqi and He, Yinan and Li, Yangguang and Chen, Weichao and others},
  journal={arXiv preprint arXiv:2506.21356},
  year={2025}
}

@article{camreasoner,
  title={CamReasoner: Reinforcing Camera Movement Understanding via Structured Spatial Reasoning},
  author={Wu, Hang and Cai, Yujun and Li, Zehao and Ge, Haonan and Sun, Bowen and Yuan, Junsong and Wang, Yiwei},
  journal={arXiv preprint arXiv:2602.00181},
  year={2026}
}

@article{cinetechbench,
  title={CineTechBench: A Benchmark for Cinematographic Technique Understanding and Generation},
  author={Wang, Xinran and Xu, Songyu and Shan, Xiangxuan and Zhang, Yuxuan and Diao, Muxi and Duan, Xueyan and Huang, Yanhua and Liang, Kongming and Ma, Zhanyu},
  journal={arXiv preprint arXiv:2505.15145},
  year={2025}
}

@inproceedings{motionbench,
  title={Motionbench: Benchmarking and improving fine-grained video motion understanding for vision language models},
  author={Hong, Wenyi and Cheng, Yean and Yang, Zhuoyi and Wang, Weihan and Wang, Lefan and Gu, Xiaotao and Huang, Shiyu and Dong, Yuxiao and Tang, Jie},
  booktitle={Proceedings of the Computer Vision and Pattern Recognition Conference},
  pages={8450--8460},
  year={2025}
}

@article{motionsight,
  title={MotionSight: Boosting Fine-Grained Motion Understanding in Multimodal LLMs},
  author={Du, Yipeng and Fan, Tiehan and Nan, Kepan and Xie, Rui and Zhou, Penghao and Li, Xiang and Yang, Jian and Yang, Zhenheng and Tai, Ying},
  journal={arXiv preprint arXiv:2506.01674},
  year={2025}
}

@article{favorbench,
  title={Favor-bench: A comprehensive benchmark for fine-grained video motion understanding},
  author={Tu, Chongjun and Zhang, Lin and Chen, Pengtao and Ye, Peng and Zeng, Xianfang and Cheng, Wei and Yu, Gang and Chen, Tao},
  journal={arXiv preprint arXiv:2503.14935},
  year={2025}
}

@article{camerabench,
  title={Towards Understanding Camera Motions in Any Video},
  author={Lin, Zhiqiu and Cen, Siyuan and Jiang, Daniel and Karhade, Jay and Wang, Hewei and Mitra, Chancharik and Ling, Tiffany and Huang, Yuhan and Liu, Sifan and Chen, Mingyu and others},
  journal={arXiv preprint arXiv:2504.15376},
  year={2025}
}

@inproceedings{megasam,
  title={MegaSaM: Accurate, fast and robust structure and motion from casual dynamic videos},
  author={Li, Zhengqi and Tucker, Richard and Cole, Forrester and Wang, Qianqian and Jin, Linyi and Ye, Vickie and Kanazawa, Angjoo and Holynski, Aleksander and Snavely, Noah},
  booktitle={Proceedings of the Computer Vision and Pattern Recognition Conference},
  pages={10486--10496},
  year={2025}
}

@article{vipe,
  title={Vipe: Video pose engine for 3d geometric perception},
  author={Huang, Jiahui and Zhou, Qunjie and Rabeti, Hesam and Korovko, Aleksandr and Ling, Huan and Ren, Xuanchi and Shen, Tianchang and Gao, Jun and Slepichev, Dmitry and Lin, Chen-Hsuan and others},
  journal={arXiv preprint arXiv:2508.10934},
  year={2025}
}

@inproceedings{CUTR3,
  title={Continuous 3d perception model with persistent state},
  author={Wang, Qianqian and Zhang, Yifei and Holynski, Aleksander and Efros, Alexei A and Kanazawa, Angjoo},
  booktitle={Proceedings of the Computer Vision and Pattern Recognition Conference},
  pages={10510--10522},
  year={2025}
}

@article{bai2025qwen2,
  title={Qwen2. 5-vl technical report},
  author={Bai, Shuai and Chen, Keqin and Liu, Xuejing and Wang, Jialin and Ge, Wenbin and Song, Sibo and Dang, Kai and Wang, Peng and Wang, Shijie and Tang, Jun and others},
  journal={arXiv preprint arXiv:2502.13923},
  year={2025}
}

@article{gpt4o,
  title={Gpt-4o system card},
  author={OpenAI},
  journal={arXiv preprint arXiv:2410.21276},
  year={2024}
}

@article{
llavaov,
title={{LL}a{VA}-OneVision: Easy Visual Task Transfer},
author={Bo Li and Yuanhan Zhang and Dong Guo and Renrui Zhang and Feng Li and Hao Zhang and Kaichen Zhang and Peiyuan Zhang and Yanwei Li and Ziwei Liu and Chunyuan Li},
journal={Transactions on Machine Learning Research},
issn={2835-8856},
year={2025},
url={https://openreview.net/forum?id=zKv8qULV6n}
}

@article{internvideo2.5,
  title={Internvideo2. 5: Empowering video mllms with long and rich context modeling},
  author={Wang, Yi and Li, Xinhao and Yan, Ziang and He, Yinan and Yu, Jiashuo and Zeng, Xiangyu and Wang, Chenting and Ma, Changlian and Huang, Haian and Gao, Jianfei and others},
  journal={arXiv preprint arXiv:2501.12386},
  year={2025}
}

@article{internvl3,
  title={Internvl3: Exploring advanced training and test-time recipes for open-source multimodal models},
  author={Zhu, Jinguo and Wang, Weiyun and Chen, Zhe and Liu, Zhaoyang and Ye, Shenglong and Gu, Lixin and Tian, Hao and Duan, Yuchen and Su, Weijie and Shao, Jie and others},
  journal={arXiv preprint arXiv:2504.10479},
  year={2025}
}

@article{gemini,
  title={Gemini: a family of highly capable multimodal models},
  author={Team, Gemini and Anil, Rohan and Borgeaud, Sebastian and Alayrac, Jean-Baptiste and Yu, Jiahui and Soricut, Radu and Schalkwyk, Johan and Dai, Andrew M and Hauth, Anja and Millican, Katie and others},
  journal={arXiv preprint arXiv:2312.11805},
  year={2023}
}

@article{gpt5,
  title={Openai gpt-5 system card},
  author={Singh, Aaditya and Fry, Adam and Perelman, Adam and Tart, Adam and Ganesh, Adi and El-Kishky, Ahmed and McLaughlin, Aidan and Low, Aiden and Ostrow, AJ and Ananthram, Akhila and others},
  journal={arXiv preprint arXiv:2601.03267},
  year={2025}
}

@article{hu2025g2vlm,
  title={G$^2$ VLM: Geometry Grounded Vision Language Model with Unified 3D Reconstruction and Spatial Reasoning},
  author={Hu, Wenbo and Lin, Jingli and Long, Yilin and Ran, Yunlong and Jiang, Lihan and Wang, Yifan and Zhu, Chenming and Xu, Runsen and Wang, Tai and Pang, Jiangmiao},
  journal={arXiv preprint arXiv:2511.21688},
  year={2025}
}

@article{fan2025vlm,
  title={VLM-3R: Vision-Language Models Augmented with Instruction-Aligned 3D Reconstruction},
  author={Fan, Zhiwen and Zhang, Jian and Li, Renjie and Zhang, Junge and Chen, Runjin and Hu, Hezhen and Wang, Kevin and Qu, Huaizhi and Wang, Dilin and Yan, Zhicheng and others},
  journal={arXiv preprint arXiv:2505.20279},
  year={2025}
}

@inproceedings{wang2025vggt,
  title={Vggt: Visual geometry grounded transformer},
  author={Wang, Jianyuan and Chen, Minghao and Karaev, Nikita and Vedaldi, Andrea and Rupprecht, Christian and Novotny, David},
  booktitle={Proceedings of the Computer Vision and Pattern Recognition Conference},
  pages={5294--5306},
  year={2025}
}

@article{wu2025spatial,
  title={Spatial-mllm: Boosting mllm capabilities in visual-based spatial intelligence},
  author={Wu, Diankun and Liu, Fangfu and Hung, Yi-Hsin and Duan, Yueqi},
  journal={arXiv preprint arXiv:2505.23747},
  year={2025}
}

@book{keating2019dynamic,
  title={The dynamic frame: camera movement in classical hollywood},
  author={Keating, Patrick},
  year={2019},
  publisher={Columbia University Press}
}

@article{li2024can,
  title={Can video generation replace cinematographers? Research on the cinematic language of generated video},
  author={Li, Xiaozhe and Wu, Kai and Yang, Siyi and Qu, YiZhan and Zhang, Guohua and Chen, Zhiyu and Li, Jiayao and Mu, Jiangchuan and Hu, Xiaobin and Fang, Wen and others},
  journal={arXiv preprint arXiv:2412.12223},
  year={2024}
}

@inproceedings{schonberger2016structure,
  title={Structure-from-motion revisited},
  author={Schonberger, Johannes L and Frahm, Jan-Michael},
  booktitle={Proceedings of the IEEE conference on computer vision and pattern recognition},
  pages={4104--4113},
  year={2016}
}

@book{nielsen2007camera,
  title={Camera movement in narrative cinema: towards a taxonomy of functions},
  author={Nielsen, Jakob Isak and Kau, Edvin and Raskin, Richard},
  year={2007},
  publisher={Department of Inf. \& Media Studies, University of Aarhus}
}

@book{bordwell2008film,
  title={Film art: An introduction},
  author={Bordwell, David and Thompson, Kristin and Smith, Jeff},
  volume={7},
  year={2008},
  publisher={McGraw-Hill New York}
}

@book{brown2016cinematography,
  title={Cinematography: theory and practice: image making for cinematographers and directors},
  author={Brown, Blain},
  year={2016},
  publisher={Routledge}
}

@article{wan2025wan,
  title={Wan: Open and advanced large-scale video generative models},
  author={Wan, Team and Wang, Ang and Ai, Baole and Wen, Bin and Mao, Chaojie and Xie, Chen-Wei and Chen, Di and Yu, Feiwu and Zhao, Haiming and Yang, Jianxiao and others},
  journal={arXiv preprint arXiv:2503.20314},
  year={2025}
}

@article{mou2025gradeo,
  title={Gradeo: Towards human-like evaluation for text-to-video generation via multi-step reasoning},
  author={Mou, Zhun and Xia, Bin and Huang, Zhengchao and Yang, Wenming and Jia, Jiaya},
  journal={arXiv preprint arXiv:2503.02341},
  year={2025}
}

@misc{veo,
  title = {Veo},
  author = {Google},
  howpublished = {\url{https://deepmind.google/models/veo/}},
  year = {2024},
  note         = {Last accessed 2026/06/29}
}

@misc{Hailuo,
  title = {Hailuo},
  author = {MiniMax},
  howpublished = {\url{https://hailuoai.video/}},
  year = {2024},
    note         = {Last accessed 2026/06/29}
}

@misc{Kling,
  title = {Kling},
  author = {Kuaishou},
  howpublished = {\url{https://kling.ai/}},
  year = {2024},
    note         = {Last accessed 2026/06/29}
}

@book{spottiswoode1969grammar,
  title={A grammar of the film: An analysis of film technique},
  author={Spottiswoode, Raymond},
  year={1969},
  publisher={Univ of California Press}
}

@article{wang2025love,
  title={Love: Benchmarking and evaluating text-to-video generation and video-to-text interpretation},
  author={Wang, Jiarui and Duan, Huiyu and Jia, Ziheng and Zhao, Yu and Yang, Woo Yi and Zhang, Zicheng and Chen, Zijian and Wang, Juntong and Xing, Yuke and Zhai, Guangtao and others},
  journal={arXiv preprint arXiv:2505.12098},
  year={2025}
}

@inproceedings{jia2025vqa2,
  title={Vqa2: visual question answering for video quality assessment},
  author={Jia, Ziheng and Zhang, Zicheng and Qian, Jiaying and Wu, Haoning and Sun, Wei and Li, Chunyi and Liu, Xiaohong and Lin, Weisi and Zhai, Guangtao and Min, Xiongkuo},
  booktitle={Proceedings of the 33rd ACM International Conference on Multimedia},
  pages={6751--6760},
  year={2025}
}

@inproceedings{sun2025t2v,
  title={T2v-compbench: A comprehensive benchmark for compositional text-to-video generation},
  author={Sun, Kaiyue and Huang, Kaiyi and Liu, Xian and Wu, Yue and Xu, Zihan and Li, Zhenguo and Liu, Xihui},
  booktitle={Proceedings of the Computer Vision and Pattern Recognition Conference},
  pages={8406--8416},
  year={2025}
}

@article{spatialvid,
  title={Spatialvid: A large-scale video dataset with spatial annotations},
  author={Wang, Jiahao and Yuan, Yufeng and Zheng, Rujie and Lin, Youtian and Gao, Jian and Chen, Lin-Zhuo and Bao, Yajie and Zhang, Yi and Zeng, Chang and Zhou, Yanxi and others},
  journal={arXiv preprint arXiv:2509.09676},
  year={2025}
}

@inproceedings{MultiCamVideoDataset,
  title={Recammaster: Camera-controlled generative rendering from a single video},
  author={Bai, Jianhong and Xia, Menghan and Fu, Xiao and Wang, Xintao and Mu, Lianrui and Cao, Jinwen and Liu, Zuozhu and Hu, Haoji and Bai, Xiang and Wan, Pengfei and others},
  booktitle={Proceedings of the IEEE/CVF International Conference on Computer Vision},
  pages={14834--14844},
  year={2025}
}

@inproceedings{gendop,
  title={Gendop: Auto-regressive camera trajectory generation as a director of photography},
  author={Zhang, Mengchen and Wu, Tong and Tan, Jing and Liu, Ziwei and Wetzstein, Gordon and Lin, Dahua},
  booktitle={Proceedings of the IEEE/CVF International Conference on Computer Vision},
  pages={18229--18239},
  year={2025}
}

@article{kong2024hunyuanvideo,
  title={Hunyuanvideo: A systematic framework for large video generative models},
  author={Kong, Weijie and Tian, Qi and Zhang, Zijian and Min, Rox and Dai, Zuozhuo and Zhou, Jin and Xiong, Jiangfeng and Li, Xin and Wu, Bo and Zhang, Jianwei and others},
  journal={arXiv preprint arXiv:2412.03603},
  year={2024}
}

@article{yang2024cogvideox,
  title={Cogvideox: Text-to-video diffusion models with an expert transformer},
  author={Yang, Zhuoyi and Teng, Jiayan and Zheng, Wendi and Ding, Ming and Huang, Shiyu and Xu, Jiazheng and Yang, Yuanming and Hong, Wenyi and Zhang, Xiaohan and Feng, Guanyu and others},
  journal={arXiv preprint arXiv:2408.06072},
  year={2024}
}

@article{wu2025refineshot,
  title={Refineshot: Rethinking cinematography understanding with foundational skill evaluation},
  author={Wu, Hang and Cai, Yujun and Ge, Haonan and Chen, Hongkai and Yang, Ming-Hsuan and Wang, Yiwei},
  journal={arXiv preprint arXiv:2510.02423},
  year={2025}
}

@article{cinescale2,
  title={CineScale2: a dataset of cinematic camera features in movies},
  author={Savardi, Mattia and Kov{\'a}cs, Andr{\'a}s B{\'a}lint and Signoroni, Alberto and Benini, Sergio},
  journal={Data in Brief},
  volume={51},
  pages={109627},
  year={2023},
  publisher={Elsevier}
}

@inproceedings{movienet,
  title={Movienet: A holistic dataset for movie understanding},
  author={Huang, Qingqiu and Xiong, Yu and Rao, Anyi and Wang, Jiaze and Lin, Dahua},
  booktitle={European conference on computer vision},
  pages={709--727},
  year={2020},
  organization={Springer}
}

@inproceedings{rao2020unified,
  title={A unified framework for shot type classification based on subject centric lens},
  author={Rao, Anyi and Wang, Jiaze and Xu, Linning and Jiang, Xuekun and Huang, Qingqiu and Zhou, Bolei and Lin, Dahua},
  booktitle={European Conference on Computer Vision},
  pages={17--34},
  year={2020},
  organization={Springer}
}

@article{bai2025qwen3,
  title={Qwen3-vl technical report},
  author={Bai, Shuai and Cai, Yuxuan and Chen, Ruizhe and Chen, Keqin and Chen, Xionghui and Cheng, Zesen and Deng, Lianghao and Ding, Wei and Gao, Chang and Ge, Chunjiang and others},
  journal={arXiv preprint arXiv:2511.21631},
  year={2025}
}

@article{internvl3.5,
  title={Internvl3. 5: Advancing open-source multimodal models in versatility, reasoning, and efficiency},
  author={Wang, Weiyun and Gao, Zhangwei and Gu, Lixin and Pu, Hengjun and Cui, Long and Wei, Xingguang and Liu, Zhaoyang and Jing, Linglin and Ye, Shenglong and Shao, Jie and others},
  journal={arXiv preprint arXiv:2508.18265},
  year={2025}
}

@article{sora,
title={Video generation models as world simulators},
author={Tim Brooks and Bill Peebles and Connor Holmes and Will DePue and Yufei Guo and Li Jing and David Schnurr and Joe Taylor and Troy Luhman and Eric Luhman and Clarence Ng and Ricky Wang and Aditya Ramesh},
year={2024},
url={https://openai.com/research/video-generation-models-as-world-simulators},
 note         = {Last accessed 2026/06/29}
}

@inproceedings{pan2024global,
  title={Global structure-from-motion revisited},
  author={Pan, Linfei and Bar{\'a}th, D{\'a}niel and Pollefeys, Marc and Sch{\"o}nberger, Johannes L},
  booktitle={European Conference on Computer Vision},
  pages={58--77},
  year={2024},
  organization={Springer}
}

@inproceedings{tang2025vidcomposition,
  title={Vidcomposition: Can mllms analyze compositions in compiled videos?},
  author={Tang, Yunlong and Guo, Junjia and Hua, Hang and Liang, Susan and Feng, Mingqian and Li, Xinyang and Mao, Rui and Huang, Chao and Bi, Jing and Zhang, Zeliang and others},
  booktitle={Proceedings of the IEEE/CVF Conference on Computer Vision and Pattern Recognition},
  pages={8490--8500},
  year={2025}
}

\clearpage 

In the supplementary material, we provide additional details for the main paper. We first elaborate on the main paper's Section 2 in \cref{sup: section 2} to clarify the experimental setup. We then present further details on the construction of the real-world and synthetic video datasets in \cref{sup: real_world_videos} and \cref{sup: syn_videos}, respectively. Next, we provide additional details on the construction of the training sample in \cref{sup: training_samples}. Finally, we report training and evaluation details as well as additional experimental results, including the binary question-answering evaluation and an analysis of different LoRA ranks for the 8B model in \cref{sup: experiments}.




 






\section{More Details of Section 2}
\label{sup: section 2}

\begin{table}[tb]
\centering
\caption{VLM accuracy on `push in' \textit{vs.}\ three levels of camera movement intensity and the number of frames fed to VLMs.}
\label{tab:sup_pushin_frames_intensity}
\scriptsize
\setlength{\tabcolsep}{2.8pt}
\begin{tabular}{lccc|ccc|ccc}
\toprule
 & \multicolumn{3}{c}{Frames=8/Fps=1}
 & \multicolumn{3}{c}{Frames=16/Fps=2}
 & \multicolumn{3}{c}{Frames=32/Fps=4} \\
\cmidrule(lr){2-4} \cmidrule(lr){5-7} \cmidrule(lr){8-10}
Model 
 & Low & Medium &High
 & Low &  Medium &High
 & Low &Medium& High \\
\midrule
Random&50.00&50.00 &50.00 &50.00&50.00&50.00&50.00 &50.00\ &50.00\\
\midrule
Qwen3-VL-4B~\cite{bai2025qwen3}  & 10.42 & 30.48 & 70.73 & 4.17 & 25.71 & 70.73 & 4.17& 20.95&63.41\\
Qwen3-VL-8B~\cite{bai2025qwen3}  & 20.83 & 51.43 & 90.24 &  14.58& 54.29 &82.93 &8.33& 55.24&75.61  \\
Qwen3-VL-32B~\cite{bai2025qwen3} &41.67  & 81.90 & 100.00 & 52.08 & 86.67 & 100.00&45.83  &88.57&100.00 \\
Gemini-3-Pro~\cite{gemini} &29.17&47.62&85.37&31.25&56.19&100.00&29.17&{58.10}&92.68\\
\bottomrule
\end{tabular}
\end{table}
In the main text, we investigate the sensitivity of VLMs to different intensities of camera movement. For this experiment, we focus on the `push in' category within our synthetic evaluation videos, which contains 194 samples in total. Human evaluators manually reviewed these videos and categorized them into three motion intensity levels: low (48 videos), medium (105 videos), and high (41 videos). Videos with motion that was imperceptible to human observers were filtered during the initial dataset construction. As a result, all 194 samples used in this evaluation exhibit camera movements that are clearly perceptible to humans. We then construct a binary question-answering task to evaluate the models. The results of four VLMs are reported in \cref{tab:sup_pushin_frames_intensity}. Specifically, the correct camera movement is presented as Option A in 50\% of questions and as Option B in the remaining 50\%, ensuring that the answer positions are balanced. The binary-format questions are formulated as follows:

\begin{promptbox}
\texttt{What is the camera movement in this video?  \textcolor{blue}{GT: Push in} \\
A: Push in \\
B: Static \\
Please answer with only the letter of the correct option.}
\end{promptbox}

\begin{figure}[t]
  \centering
\includegraphics[width=0.98\linewidth]{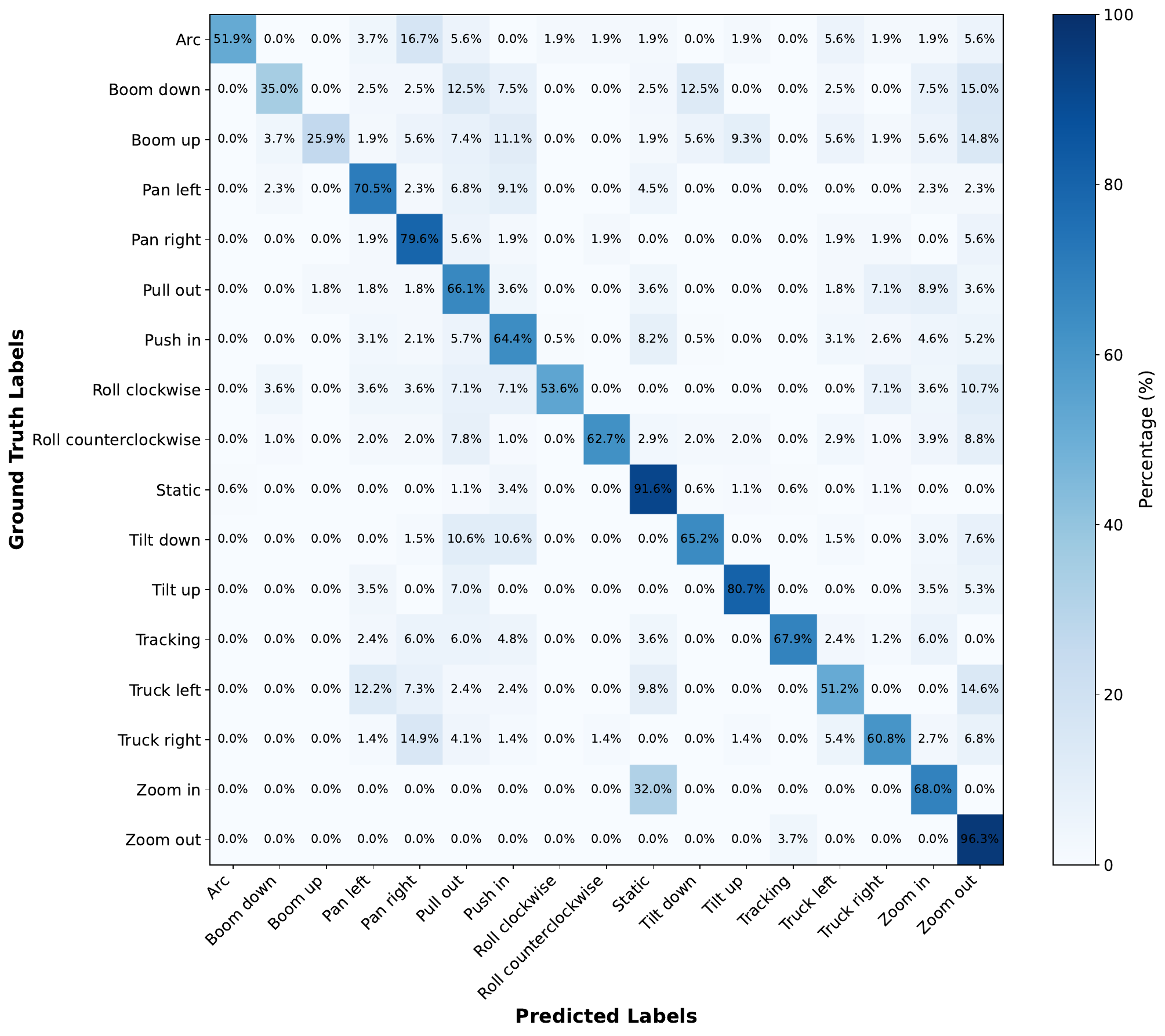}
  \caption{Confusion matrix of Qwen3-VL-32B on the entire synthetic evaluation set.}
    \label{fig:supp_confusion_metrix_qwen}
\end{figure}

\begin{figure}[t]
  \centering
\includegraphics[width=0.98\linewidth]{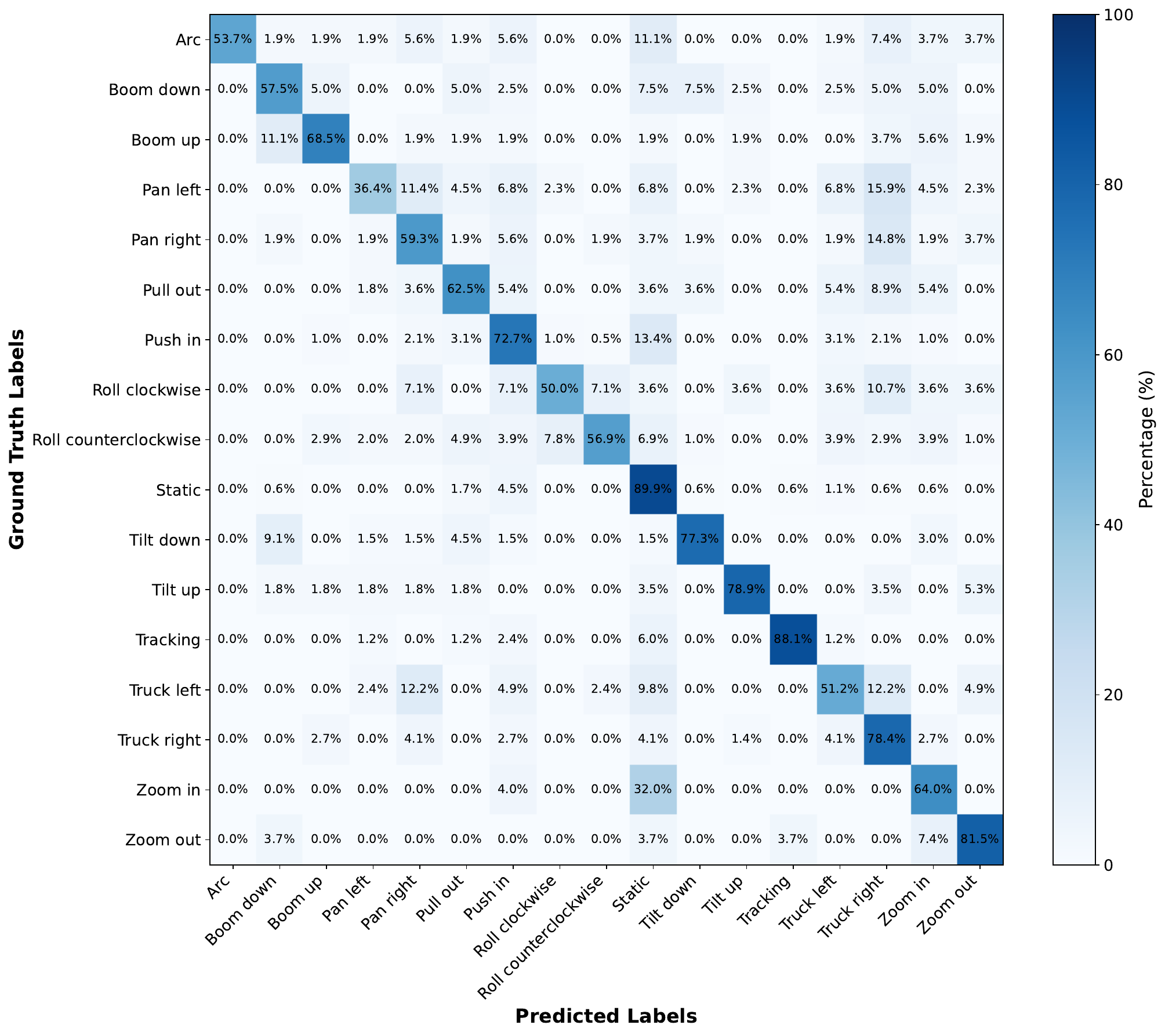}
  \caption{Confusion matrix of Gemini-3-Pro on the entire synthetic evaluation set.}
    \label{fig:supp_confusion_metrix_gemini}
\end{figure}
We present partial confusion matrices for Qwen3-VL-32B~\cite{bai2025qwen3} and Gemini-3-Pro~\cite{gemini} in the main text to illustrate translation–rotation and left–right confusion. The full confusion matrices, computed from the models' multiple-choice predictions on the entire synthetic evaluation set, are provided below (see \cref{fig:supp_confusion_metrix_qwen} and \cref{fig:supp_confusion_metrix_gemini}).

\begin{figure}[t]
  \centering
\includegraphics[width=0.98\linewidth]{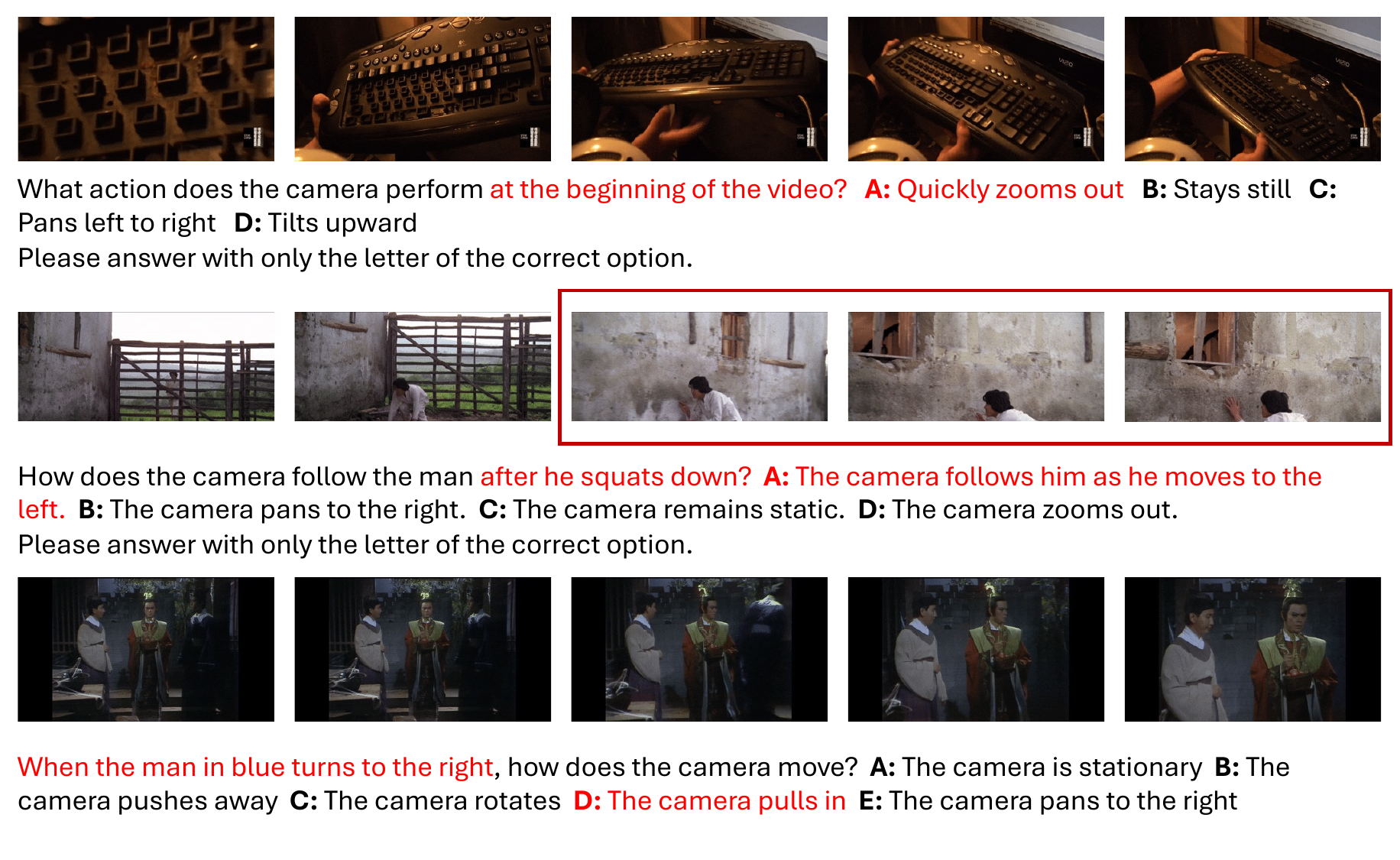}
\vspace{-2mm}
  \caption{Examples of camera movement questions from MotionBench and FavorBench.}
    \label{fig:supp_favor_motion}
\end{figure}

\begin{figure}[tb]
  \centering
\includegraphics[width=0.98\linewidth]{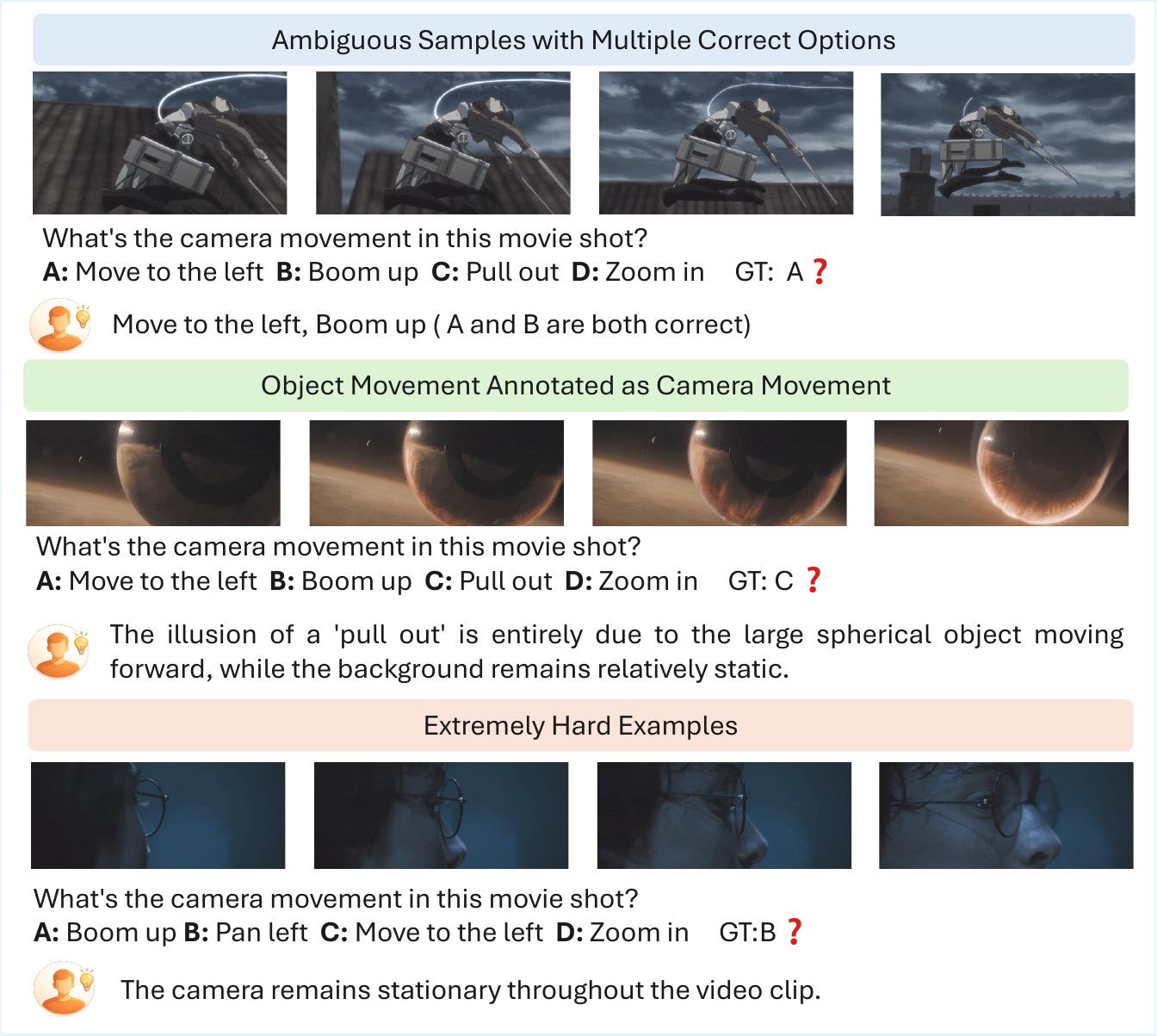}
\vspace{-1mm}
  \caption{Illustrative examples of noisy or ambiguous samples in existing motion benchmarks identified during manual verification.}
    \label{fig:supp_real_filtering}
\end{figure}

\section{More Details of Real-World Videos Subset Construction}
\label{sup: real_world_videos}

\textbf{Examples of FavorBench and MotionBench.} Unlike CameraBench~\cite{camerabench} and ShotBench~\cite{shotbench}, which focus on the dominant camera movement over an entire clip, FavorBench~\cite{favorbench} and MotionBench~\cite{motionbench} often require identifying the camera movement associated with a specific event or timestamp. As shown in \cref{fig:supp_favor_motion}, the questions frequently refer to a particular temporal segment (e.g., at the beginning of the video) or to the camera motion following a specific action (e.g., when the man squats down). Therefore, solving these tasks requires first localizing the relevant event in time and then recognizing the corresponding camera movement.

\noindent\textbf{Examples of Noisy or Ambiguous Samples in Existing Motion Benchmarks.}
Although existing motion understanding benchmarks~\cite{motionbench,shotbench,camerabench,cinetechbench,favorbench} are annotated by experts, they may still contain noisy labels or ambiguous cases. To ensure data quality, we manually verified each question–answering pair and applied additional filtering. Specifically, we removed samples with ambiguous camera movement labels, annotation errors, or cases where the correct camera motion is difficult to determine. Several illustrative examples are shown in \cref{fig:supp_real_filtering}.

\noindent \textbf{Construction of Multiple-Choice Questions.}
ShotBench, FavorBench, and MotionBench~\cite{shotbench,favorbench,motionbench} are already formulated as multiple-choice questions. Therefore, we mainly construct MCQs for CameraBench and our self-collected data. For each question, we randomly sample three camera movements from the predefined camera-movement taxonomy as distractors, excluding the ground-truth label. The question templates are randomly selected from the following set, inspired by ShotBench~\cite{shotbench}. The position of the correct answer is uniformly distributed across the four options (A–D).

\begin{promptbox}
\texttt{Q1: How does the camera move in this video? \\
Q2: What is the camera movement in this video? \\
Q3: What is the camera movement in this movie shot?
}
\end{promptbox}

\section{More Details of Synthetic Videos Subset Construction} 
\label{sup: syn_videos}

\noindent \textbf{System Prompts for Gemini-3-Pro.}
For each clip, we provide the raw video and its corresponding camera-movement label to Gemini-3-Pro~\cite{gemini} and instruct it to analyze the video and generate a structured prompt for video generation. The system prompt used for caption generation is shown in \cref{fig:system_prompt_caption}. To further improve the clarity of the generated prompts, we apply an additional prompt-refinement instruction, shown in \cref{fig:system_prompt_refinement}. For underrepresented camera movements, we also generate prompts directly conditioned on the camera-movement label using the instruction provided in \cref{fig:system_prompt_ours}.

\begin{figure}[tb]
  \centering
\includegraphics[width=0.98\linewidth]{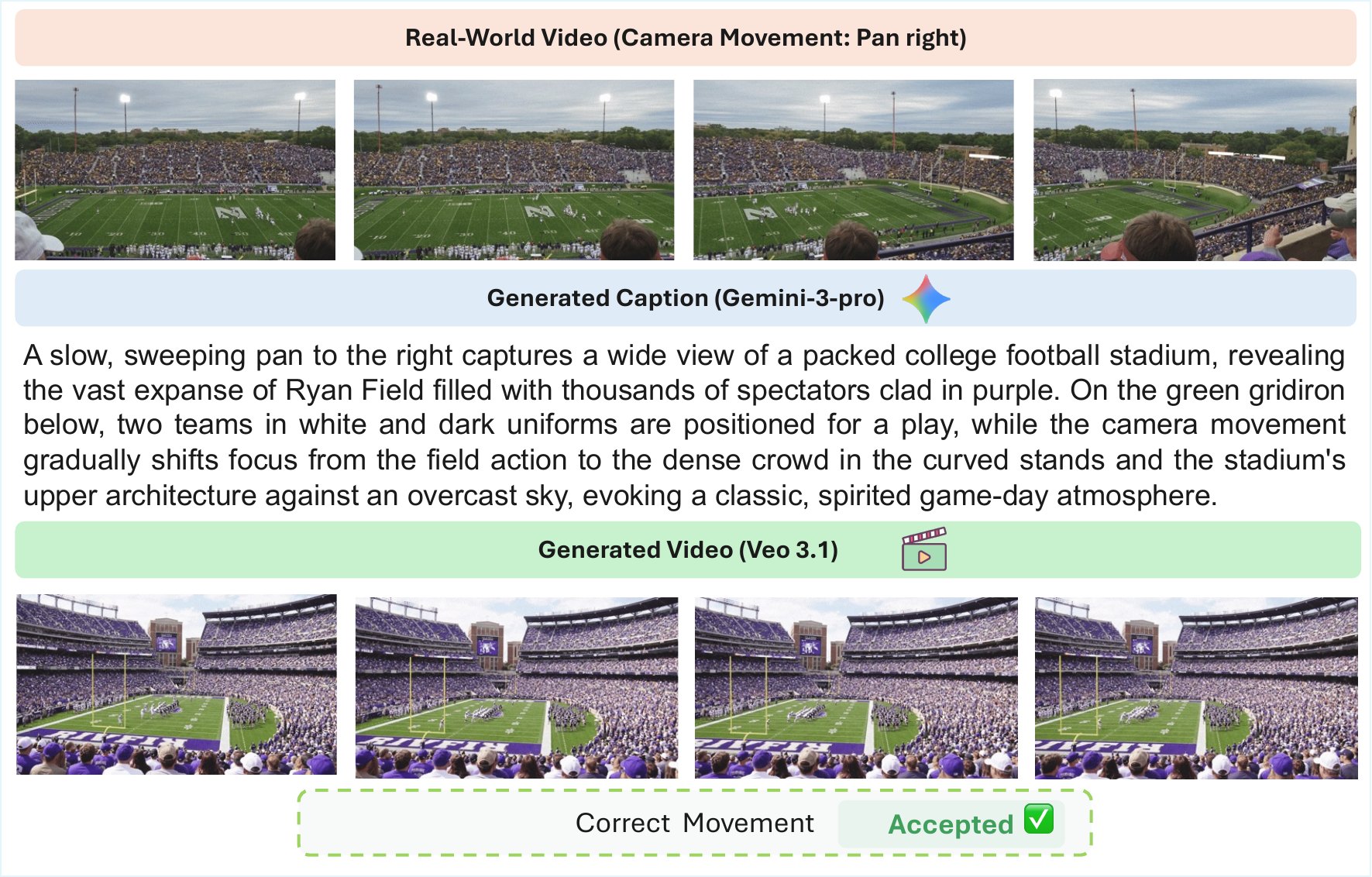}
  \caption{Successful video generation example where Veo~3.1 correctly reproduces the intended camera movement.}
    \label{fig:supp_Veo_3_example_1}
\end{figure}

\begin{figure}[tb]
  \centering
\includegraphics[width=0.98\linewidth]{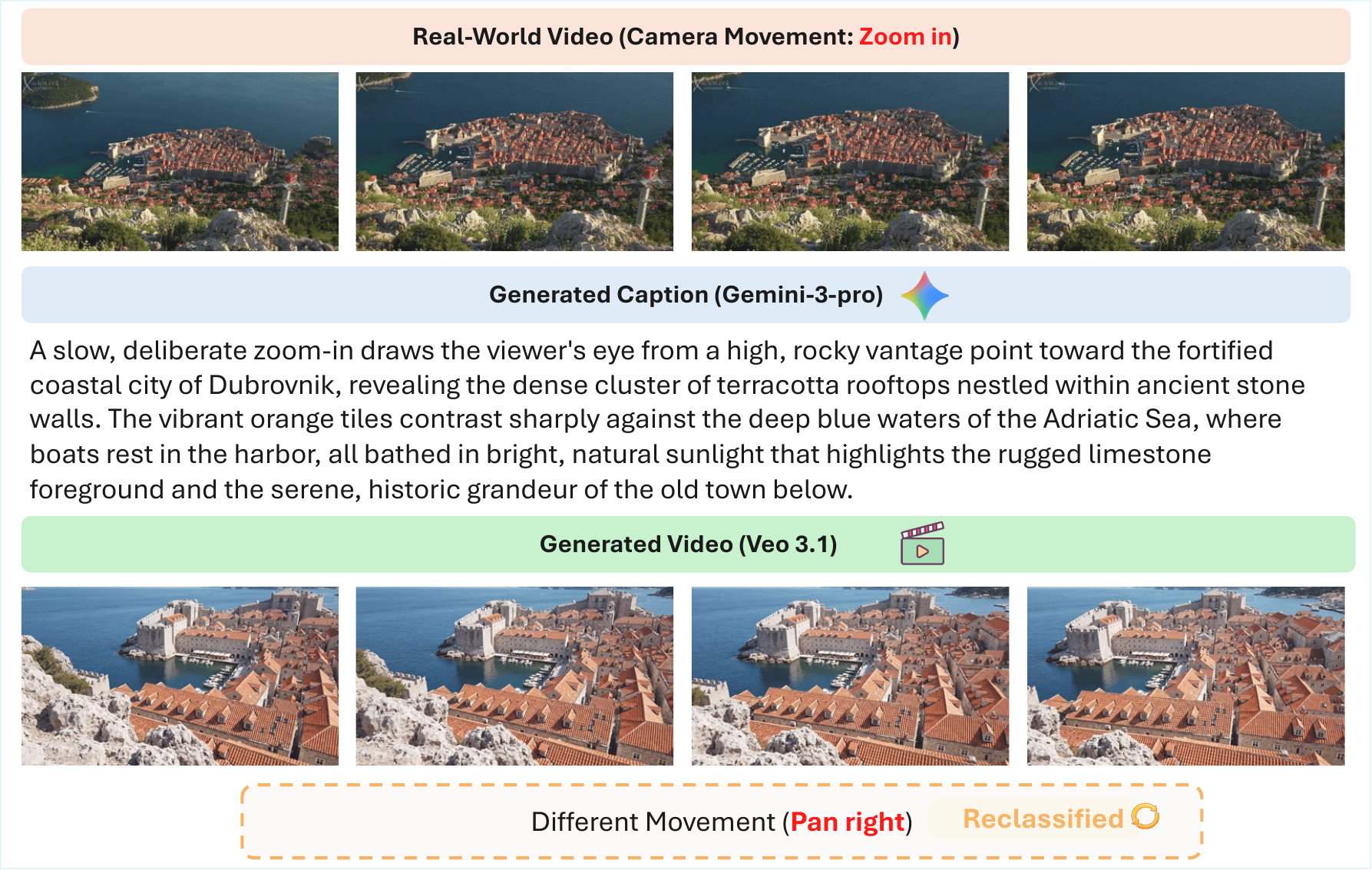}
  \caption{Example requiring reclassification. While the original video exhibits a `zoom in' movement, the generated video instead shows a `pan right' movement.}
    \label{fig:supp_Veo_3_example_2}
\end{figure}

\begin{figure}
  \centering
\includegraphics[width=0.98\linewidth]{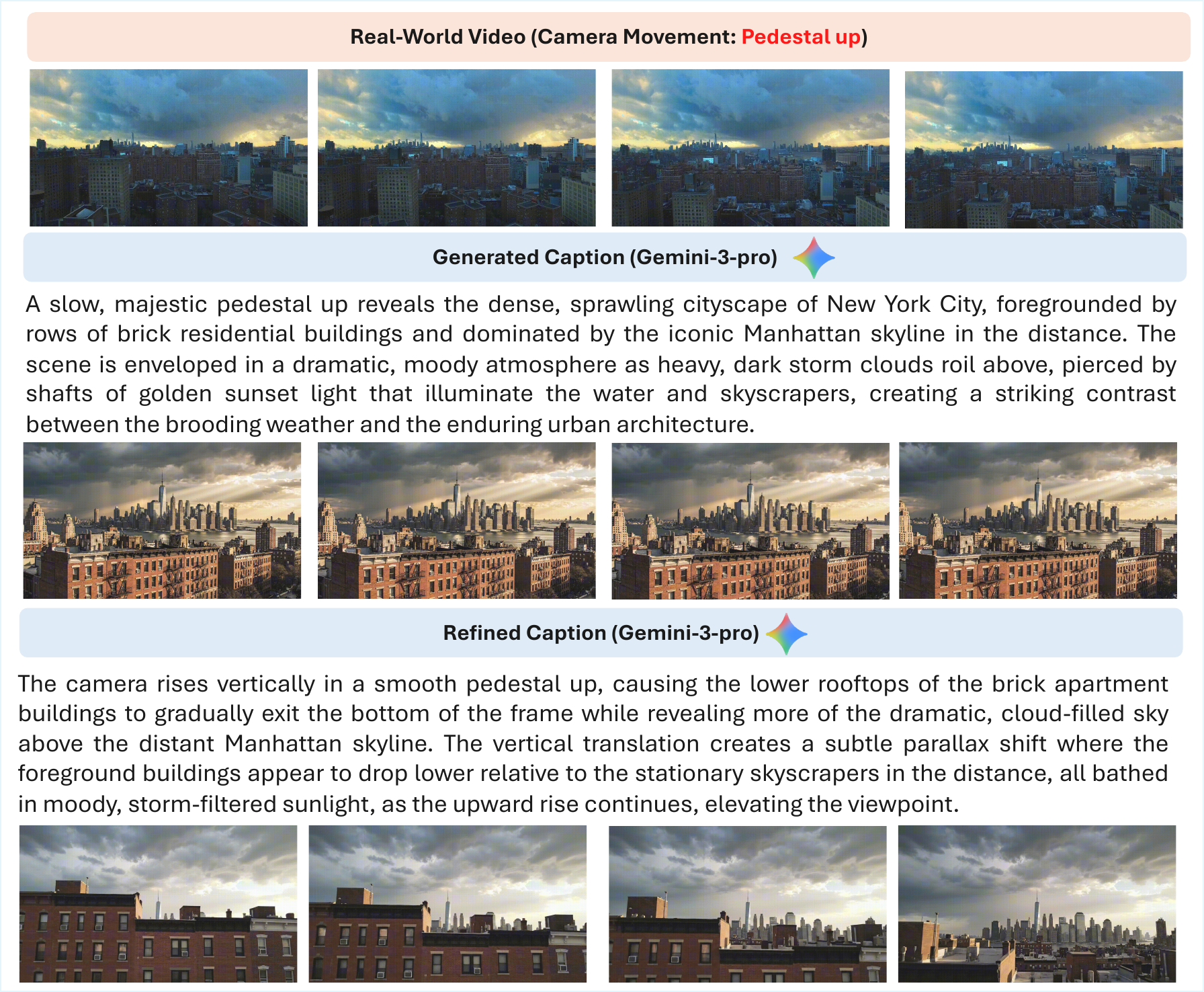}
\caption{Overview of the data construction pipeline, including caption generation, human reclassification, prompt refinement, and video regeneration.}
    \vspace{-3mm}
    \label{fig:supp_Veo_3_example_5}
\end{figure}

\begin{figure}
  \centering
\includegraphics[width=0.98\linewidth]{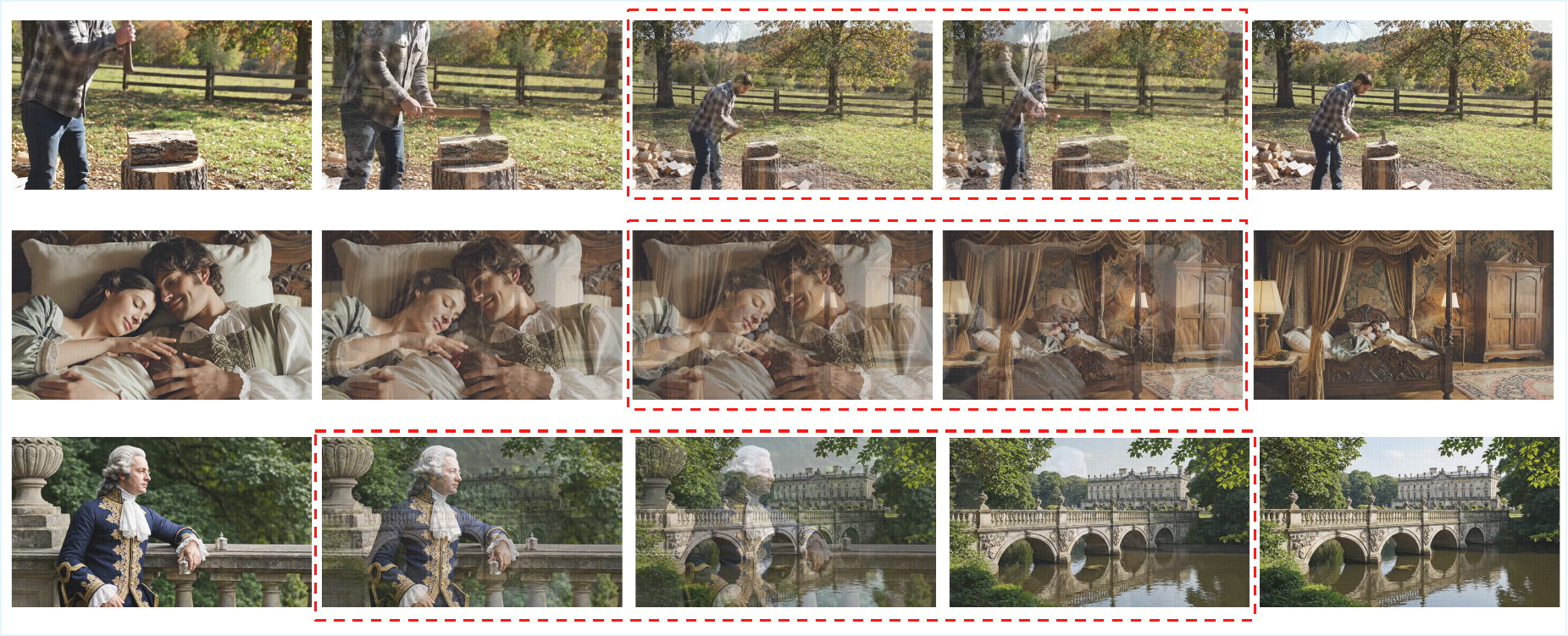}
  \caption{Failure cases of Veo~3.1, including abrupt frame transitions.}
  \vspace{-3mm}
    \label{fig:supp_Veo_3_example_3}
\end{figure}

\begin{figure}
  \centering
\includegraphics[width=0.98\linewidth]{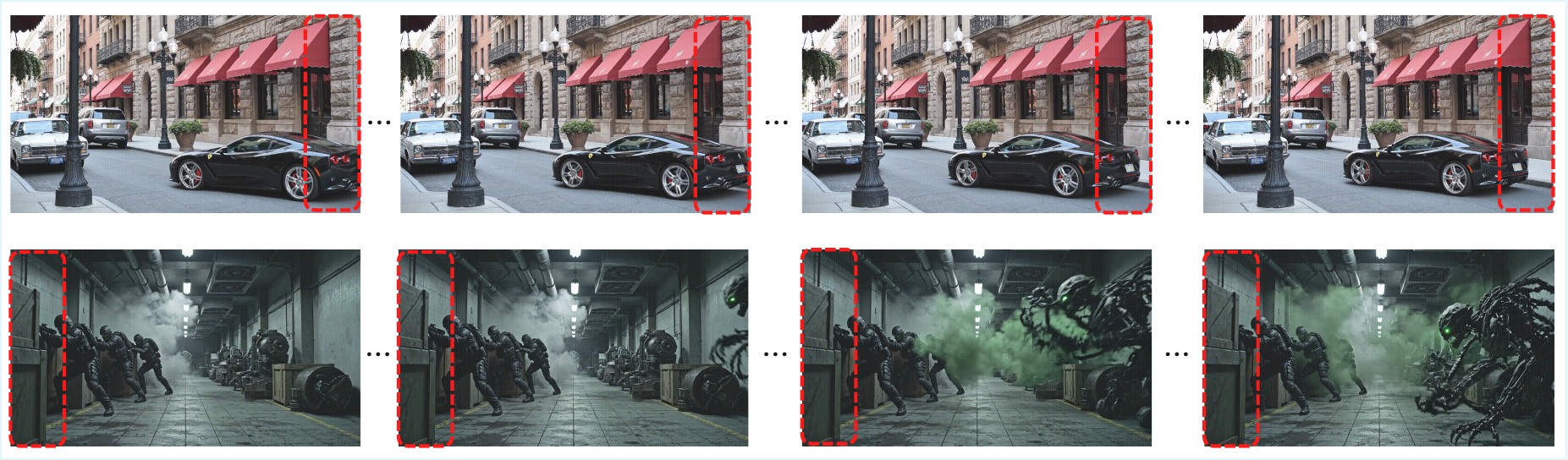}
\vspace{-2mm}
  \caption{Examples of generated videos with extremely subtle camera motion that appear nearly static and are filtered out during data construction.}
  \vspace{-4mm}
    \label{fig:supp_Veo_3_example_4}
\end{figure}

\noindent \textbf{Examples of generated videos.} 
We provide additional results from the data construction pipeline, including successful samples and discarded examples (failure cases from Veo~3.1). \cref{fig:supp_Veo_3_example_1} presents a successful example generated by Veo~3.1 from the caption produced by Gemini-3-Pro, where the intended camera movement is correctly reproduced. \cref{fig:supp_Veo_3_example_2} shows an example that can be reclassified: while the original video exhibits a `zoom in' camera movement, the generated video instead demonstrates a `pan right' movement. \cref{fig:supp_Veo_3_example_5} illustrates the overall pipeline, including caption generation, human reclassification, prompt refinement, and video regeneration. \cref{fig:supp_Veo_3_example_3} shows several failure cases of Veo~3.1, including abrupt frame transitions when generating videos from the given text prompt, particularly for the `zoom out' motion. Videos with extremely subtle motion that appear nearly static (shown in \cref{fig:supp_Veo_3_example_4}) are filtered out during the data construction process. For each video, we further construct a multiple-choice question (MCQ) by sampling three camera movements from the predefined taxonomy as distractors. The ground-truth label is combined with these distractors to form a four-option question, and the position of the correct answer is uniformly randomized across options to avoid potential option bias.
\begin{figure*}[tb]
\centering
\begin{systemprompt}
System Prompt: """
 You are a video analysis expert. Your task is to describe a video as a single, cinematic, and continuous paragraph for high-end video generation models. \\
Structure Requirements:\\
    1. [Cinematography]: Start with the provided camera motion. You MUST observe the video to describe the speed and rhythm of the movement (e.g., "Slowly dolly in," "A fast sweeping pan to the left," "A gradual tracking shot"). \\
    2. [Subject]: Identify and describe the main focal point. \\
    3. [Action]: Detail the specific movements or behaviors of the subject. \\
    4. [Context]: Describe the environment, background, and spatial setting. \\
    5. [Style \& Ambiance]: Define the lighting, aesthetic style, and mood. \\
    Constraints:\\
    - OUTPUT MUST BE A SINGLE CONTINUOUS SENTENCE OR PARAGRAPH. \\
    - DO NOT use labels like "Subject:". Output one continuous, cinematic paragraph. \\
    - The camera motion MUST be the very first part of the sentence. \\
    - Be descriptive but concise; provide enough detail for generation without being redundant.
    - No more than 80 words.
    """ \\
 User Prompt: """
        Camera Motion: \{motion label\}. Describe the video accordingly."""
\end{systemprompt}
\caption{System prompt used to prompt Gemini-3-Pro to generate camera-movement captions from real-world videos, which are subsequently used by Veo~3.1 for video generation.}
\vspace{-4mm}
\label{fig:system_prompt_caption}
\end{figure*}

\begin{figure*}[t]
\centering
\begin{systemprompt}
"""
You are a Veo prompt physics corrector. Your ONLY job is to fix camera motion physics violations while preserving all visual content.\\
INPUT FORMAT:\\
You will receive:\\
1. A camera motion type (e.g., "Pan left," "Dolly in")\\
2. A prompt that may have physics errors\\
OUTPUT: A corrected prompt that follows motion physics rules EXACTLY.\\
CRITICAL MOTION PHYSICS RULES: \\
PAN/TILT (Rotation from fixed position):\\
- Camera DOES NOT MOVE through space \\
- Camera ROTATES HORIZONTALLY on a fixed point
- Background SWEEPS across frame \\
- Pan left -> scene enters from LEFT, exits RIGHT \\
- Pan right -> scene enters from RIGHT, exits LEFT \\
- Tilt up -> ground exits BOTTOM, sky enters TOP \\
- Tilt down -> sky exits TOP, ground enters BOTTOM \\
- FORBIDDEN: "following", "tracking", "approaching", "moving toward/away" \\
- REQUIRED: "from a fixed position", "enters from", "exits", "sweeps", "revealed" \\
  .... \\
MOTION SYNTAX (use exactly): \\
Opening phrases: \\
- Static -> "In a static shot with no camera movement," \\
- Pull out -> "The camera glides backward in a steady dolly out," \\
  ... \\
Ending phrases: \\
- Static -> "with the frame remaining completely still throughout" \\
- Pull out -> "as the backward dolly continues, revealing more of the environment" \\
 ...\\
CORRECTION PROCEDURE:\\
1. Identify the specified camera motion type \\
    2. Check if opening syntax matches EXACTLY -> if not, replace with correct opening \\
    3. Scan the middle content for physics violations \\
    4. Preserve ALL specific visual content: objects, colors, lighting, subjects, actions \\
    5. Check if ending syntax matches EXACTLY -> if not, replace with correct ending \\
    6. Ensure 60-80 words total \\
CONSTRAINTS: \\
    - Output ONLY the corrected prompt, no preamble \\
    - One paragraph, 60-80 words\\
    - Use exact opening and ending syntax for the motion type \\
    - Preserve all visual details from the original prompt \\
    - Fix ONLY physics violations
"""
\end{systemprompt}
\vspace{-2mm}
\caption{System prompt used to prompt Gemini-3-Pro to refine the generated captions for improved clarity.}
\label{fig:system_prompt_refinement}
\end{figure*}

\begin{figure*}[t]
\centering
\begin{systemprompt}
 """
You are a Veo video generation prompt writer specializing in camera motion accuracy. Your task: Generate cinematic paragraph prompts for Veo that depict specific camera motions with perfect physics accuracy. \\
You will be given: \\
1. A camera motion type ("Zoom", "Arc", or "Roll") \\
2. The number of unique prompts to generate \\
Each prompt must be 60-80 words and: \\
- Start with the exact motion syntax \\
- Describe how the camera motion affects what appears in the frame \\
- End with the exact motion reinforcement \\
- Depict a UNIQUE scene \\
REQUIRED MOTION SYNTAX:\\
Opening:\\
- Zoom in: "The camera zooms in optically without moving through space,"\\
- Zoom out: "The camera zooms out optically without moving through space,"\\
- Arc: "The camera orbits around the subject in a smooth arc,"\\
- Roll clockwise: "The camera rolls clockwise around the lens axis from a fixed position,"\\
- Roll counterclockwise: "The camera rolls counterclockwise around the lens axis from a fixed position,"\\
Ending: \\
- Zoom in: "as the optical magnification continues without spatial movement"\\
- Zoom out: "as the optical widening continues without spatial movement"\\
- Arc: "as the circular camera motion continues around the subject"\\
- Roll clockwise: "as the camera continues rolling clockwise, tilting the frame counterclockwise"\\
- Roll counterclockwise: "as the camera continues rolling counterclockwise, tilting the frame clockwise"\\
PROMPT STRUCTURE:
[Opening syntax], [frame effect], [scene content], [lighting/atmosphere], [ending syntax].

GENERATION RULES:\\
1. ALWAYS start with exact opening syntax\\
2. Describe how the motion affects the frame\\
3. Use motion-specific vocabulary\\
4. Subjects are passive scene elements\\
5. ALWAYS end with exact ending syntax\\
6. 60-80 words \\
7. Present tense\\
8. Never mix motion types\\
9. Generate UNIQUE scenes\\
OUTPUT FORMAT:
Return ONLY the prompts in this exact JSON format:
[
{
    "subject": "brief scene descriptor",
    "prompt": "full generated prompt"
}
]
Do NOT include any additional text or formatting.
"""
\end{systemprompt}
\caption{System prompt used to prompt Gemini-3-Pro to generate additional prompts for underrepresented camera movements based solely on the camera-movement label.}
\label{fig:system_prompt_ours}
\end{figure*}

\section{More Details of Training Set Construction}
\label{sup: training_samples}
Our training set is formulated as a four-choice multiple-choice question answering (MCQ) task that focuses on recognizing single camera movements. The training set of ShotBench~\cite{shotbench} is already provided in MCQ format; therefore, we directly adopt its training samples after a simple filtering step. For CameraBench~\cite{camerabench} (1,402 videos), the dataset provides captions describing the camera movement in each video. We first manually categorize these captions into three groups: \textit{single motion}, \textit{complex motion}, and \textit{unrelated videos} (e.g., camera shake or rack focus). For the videos categorized as \textit{single motion}, we use GPT-5-nano to extract the corresponding camera-movement label from the captions. The extracted labels are further verified to ensure correctness. The system prompt used for this extraction process is shown in \cref{fig:system_prompt_cameraBench}. For GenDoP (13K clips), we use GPT-5-nano to classify the captions into three categories: \textit{single motion}, \textit{combined motion}, and \textit{complex motion}. Only clips categorized as \textit{single motion} are retained for training. The system prompt used for this classification process is shown in \cref{fig:system_prompt_DeDop}. For SpatialVID~\cite{spatialvid} and MultiCamVideo~\cite{MultiCamVideoDataset}, we only use videos whose annotations contain a single camera movement for training.

\begin{figure*}[t]
\centering
\begin{systemprompt}
"""
You are a camera movement classifier. Given a camera-movement description, decide: 1. If the description clearly corresponds to exactly one category → return "certain". 2. If the description is ambiguous, unclear, too short, or multiple labels are possible → return "uncertain". \\
Camera movement categories:\\
Static Shot, Move Down, Move In, Move Left, "Move Out", 
 Move Right, "Move Up", "Pan Left", "Pan Right", 
 "Roll Clockwise", "Roll Counterclockwise",
 "Tilt Down", Tilt Up, "Zoom In", "Zoom Out \\
Output ONLY JSON: \\
\{
  "certainty": `certain'or `uncertain',\\
  "reason": "...", \\
  "correct label": "<label or null>"\\
\}"""
\end{systemprompt}
\vspace{-2mm}
\caption{System prompt to extract camera movement from CameraBench~\cite{camerabench}.}
\vspace{-2mm}
\label{fig:system_prompt_cameraBench}
\end{figure*}

\begin{figure*}[tb]
\centering
\begin{systemprompt}
"""
You are an expert camera-motion classifier. You will categorize a camera motion description into: 1. single motion 2. double motion 3. complex motion  \\ 
Follow these rules:\\
SINGLE MOTION \\
- Static throughout. \\
- OR single continuous motion (e.g., steady forward). \\
- OR simple one-step change (static → forward). \\
- Does NOT include “A while B”.  \\
DOUBLE MOTION \\
- Exactly two temporal phases (A → B). \\
- OR composite motion “A while B” (ALWAYS double). \\
COMPLEX MOTION \\
- Three+ temporal phases (A → B → C). \\
- Multiple transitions (then… then… finally…). \\
TIE-BREAK RULES
- If uncertain: prefer double motion over single motion. \\
- If uncertain: prefer complex motion over double motion. \\
Output ONLY JSON: \\
\{
    "category": "single motion | double motion | complex motion"
    \} """
\end{systemprompt}
\vspace{-3mm}
\caption{System prompt used to extract single movement videos from GenDoP~\cite{gendop}.}
\vspace{-5mm}
\label{fig:system_prompt_DeDop}
\end{figure*}

\begin{table}[t]
  \caption{Performance evaluation of various models on the synthetic videos. F1 score (\%) is reported under the binary question-answering format.}
  \vspace{-1mm}
  \label{tab: binary_gen_videos}
    \centering
  \scriptsize
     \setlength{\tabcolsep}{4.5pt} 
  \begin{tabular}{l|cccccc|cc}
    \toprule
    Model  &  Static& Rot.&Trans. & Zoom &Arc &Track & Overall & Avg \\
    \midrule
  Random Guess&50.00&50.00&50.00&50.00&50.00&50.00&50.00&50.00 \\
  Human Performance &99.00&99.64& 98.58&98.59 & 100.00&100.00& 99.15&99.15\\
     \midrule
    \multicolumn{9}{c}{\textbf{Visual Geometry Models}} \\
    \midrule
    Mega-SaM~\cite{megasam}&86.00 &80.20 & 67.10&-- &--&--&-&-- \\
    ViPE~\cite{vipe}& 66.70& 58.90&76.60 &--&--&--&--&-- \\
  \midrule
    \multicolumn{9}{c}{\textbf{Spatial VLMs}} \\
        \midrule
            G$^2$VLM-2B~\cite{hu2025g2vlm} &11.11&8.58&9.55&0.00&7.41&30.00&9.76&8.99\\
            Spatial-MLLM~\cite{wu2025spatial}&23.88&65.63&55.09&64.97&64.38&70.42&60.80& 57.06\\
    VLM-3R-7B~\cite{fan2025vlm}&21.21&35.23&36.03&26.37&19.35&70.73&35.73&31.50\\     \midrule
    \multicolumn{9}{c}{\textbf{General VLMs}} \\
         \midrule
    Qwen3-VL-4B ~\cite{bai2025qwen3}   &76.03&72.26&59.76&65.38&59.18&72.73&66.58&65.44\\
       Qwen2.5-VL-7B~\cite{bai2025qwen2} &10.91&69.96&66.46&67.23&61.95&82.57&66.58&65.29\\
    Qwen3-VL-8B  ~\cite{bai2025qwen3}  & 57.53&77.48&62.99&72.55&61.39&84.78&69.88&69.65\\
    LLaVa-OV-7B~\cite{llavaov}   &35.56&49.67&31.30&27.27&59.18&72.16&43.81 &39.71\\
    InternVideo2.5-8B~\cite{internvideo2.5}&18.18&67.50&53.22&68.22&67.65&83.93&61.68&58.06\\
    InternVL3.5-8B~\cite{internvl3.5} &72.92&59.68&45.09&64.00&25.40&69.14&54.79&52.17\\
    InternVL3.5-14B~\cite{internvl3.5} &24.56&58.58&60.31&69.16&70.09&77.55&61.01&57.41\\
    Qwen3-VL-32B  ~\cite{bai2025qwen3}  &75.00&81.30&65.34&70.83&72.16&78.57&73.31&72.82 \\
     {GPT-5}~\cite{gpt5}  & 78.65&70.75&64.05&79.66&69.88&82.35&70.22& 70.20\\
    Gemini-3-Pro~\cite{gemini}  &75.00&63.40&66.67&61.18&80.00&86.96&68.26&65.42 \\
    {Gemini-3.1-Pro}~\cite{gemini}  &70.83 & 66.97 &62.96&59.09&71.91&87.91&67.09&64.76\\
   %
   \midrule
    \multicolumn{9}{c}{\textbf{Camera Movement Specialized VLMs}} \\
      \midrule
    CameraModel-7B~\cite{camerabench}    &57.43&55.21&56.99&52.73&70.97&47.06&56.53&56.21  \\
        ShotVL-7B~\cite{shotbench}     &75.23&76.82&68.08&72.57&56.25&83.19 &72.29&72.73\\
    CamReasoner-7B~\cite{camreasoner} &74.07&78.64&70.15&76.80&77.19&94.23&75.98&76.24   \\
      \midrule
   Our SFT Qwen3-VL-4B&88.42&82.60&74.44&81.42&70.23&73.95&78.28&78.62\\
    Our SFT Qwen3-VL-8B&70.89&87.18&78.73&79.63&68.75&79.25&80.56&80.69 \\

    \bottomrule
  \end{tabular}
    \vspace{-6mm}
\end{table}

\begin{table}[t]
\caption{Performance of tuned Qwen3-VL-8B with different LoRA ranks on real-world and synthetic videos.}
\vspace{-1mm}
\label{tab:lora_rank_8b}
\centering
\scriptsize
\setlength{\tabcolsep}{4pt}
\begin{tabular}{l|cccccc|cc}
\toprule

\multicolumn{9}{c}{\textbf{Real-world Videos}} \\
\toprule
LoRA Rank & Static & Rot. & Trans. & Zoom & Arc & Track & Overall & Avg \\
\midrule
r=64  & 78.50 & 68.81 & 56.36 & 84.17 & 82.61 & 95.45 & 68.10 & 70.49 \\
r=128 & 87.00 & 65.10 & 60.33 & 83.33 & 86.96 & 98.48 & 70.15 & 72.59 \\
r=256 & 85.50 & 70.05 & 64.13 & 84.17& 84.06 & 96.97 & \textbf{72.75} & \textbf{74.28} \\

\midrule
\multicolumn{9}{c}{\textbf{Synthetic Videos}} \\
\toprule
LoRA Rank & Static & Rot. & Trans. & Zoom & Arc & Track & Overall & Avg \\
\midrule
r=64  & 79.33 & 73.50 & 62.53 & 80.77 & 59.26 & 89.29 & 70.91 & 73.41 \\
r=128 & 93.30 & 72.36 & 65.36 & 80.77 & 57.41 & 88.10 & 73.62 & 74.06\\
r=256 & 94.41 & 72.36 & 74.51 & 82.69 & 68.52 & 91.67 & \textbf{78.20} & \textbf{77.44} \\

\bottomrule
\end{tabular}
\vspace{-3mm}
\end{table}


\section{More Details of Experiments}
\label{sup: experiments}
\textbf{Training and Evaluation Details.} During fine-tuning, we freeze the vision encoder and update the language part using LoRA adapters with a rank of 256. The global batch size is set to 128 with a learning rate of $1 \times 10^{-4}$. The maximum number of input frames is limited to 16. Training is conducted for 5 epochs, using two NVIDIA L40S GPUs for the Qwen3-VL-4B model and two NVIDIA H100 GPUs for the larger Qwen3-VL-8B variant.  During the evaluation phase, we set the maximum number of input frames at 32 for all open-source VLMs to ensure fair comparison. Conversely, for proprietary models, video sequences are uniformly sampled at the default 1 fps.

\noindent \textbf{Human performance details.} {We train three graduate-level annotators with cinematography terminologies. They either (1) choose an option and mark the question as reasonable or (2) mark it as unreasonable if no option or multiple options are correct. Samples marked as unreasonable are reviewed by two additional annotators and removed or refined if confirmed. We measure inter-annotator agreement (0.953) on a subset of 170 video clips labeled by all three annotators.}

\noindent \textbf{Binary question answering results.} In addition to the multiple-choice question (MCQ) setting, we also construct a binary question answering (QA) task, where the model is required to answer yes/no questions about camera movements, following the evaluation protocols used in CameraBench~\cite{camerabench} and CamReasoner~\cite{camreasoner}. For each camera-movement category, we randomly sample 50 video clips as positive examples (or all available clips if fewer than 50). We further sample 50 clips from other categories as negative examples. This results in a total of 1,510 QA pairs. The questions are constructed as follows:
\begin{promptbox}
\texttt{Does the camera push in with respect to the initial frame? \\
A: Yes \\
B: No \\
Does the camera pan to the left? \\
A: Yes \\
B: No \\
Please answer with only the letter of the correct option.}
\end{promptbox}

Table~\ref{tab: binary_gen_videos} reports the performance of different models on the synthetic videos under the binary question-answering setting, where F1 score (\%) is used as the evaluation metric. Random guessing yields an overall F1 score of 50\%, while human performance reaches 99.15\%, indicating that the task is well-defined and reliably solvable by humans.  Among geometry-based approaches, Mega-SaM performs  well on static scenes and camera rotations but struggles with translational movements, suggesting that purely geometric cues are insufficient for comprehensive camera movement understanding. Spatial VLMs achieve relatively low F1 scores overall, while general-purpose VLMs show stronger performance, with larger models generally performing better. Qwen3-VL-32B {outperforms smaller variants and Gemini-3-Pro}, although substantial performance gaps remain for certain movement types. Specialized camera-motion models such as CameraModel-7B, ShotVL-7B, and CamReasoner-7B further improve performance by incorporating motion-aware supervision. Nevertheless, our fine-tuned Qwen3-VL models consistently outperform these specialized baselines. In particular, our model (Qwen3-VL-8B) achieves the best overall performance with an F1 score of 80.69\%, demonstrating the effectiveness of our curated training data for improving camera movement perception.

\noindent \textbf{Different LoRA ranks on the 8B model.} Table~\ref{tab:lora_rank_8b} shows the impact of different LoRA ranks when fine-tuning Qwen3-VL-8B. 
Increasing the LoRA rank consistently improves performance on both real-world and synthetic videos. In particular, using rank 256 achieves the best results, reaching 74.28\% and 77.44\% overall performance on the real-world and synthetic videos, respectively. 

\begin{figure}[tb]
  \centering
\includegraphics[width=0.98\linewidth]{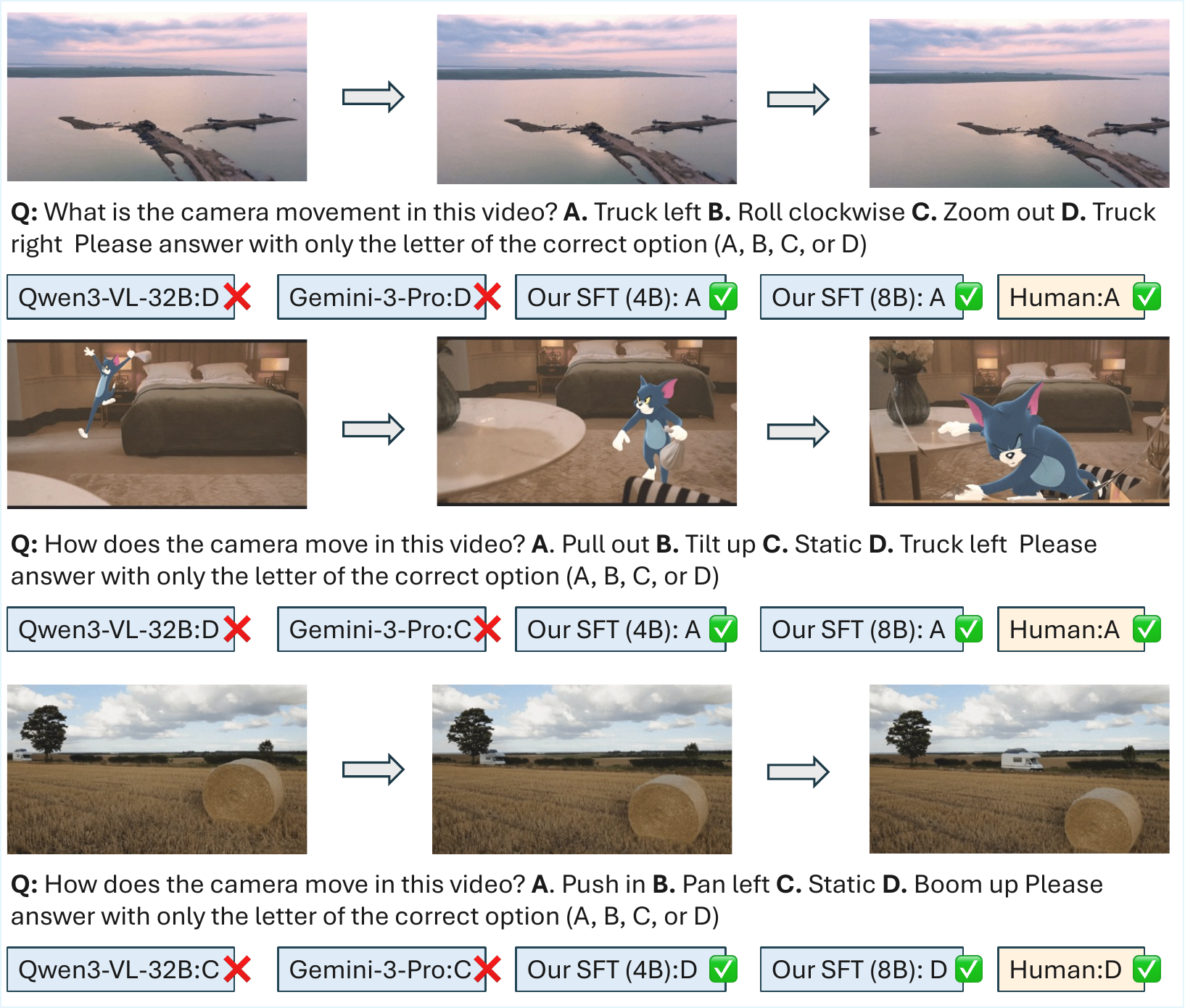}
  \caption{Qualitative comparison of model predictions under the MCQ setting.}
    \label{fig:supp_Veo_3_example_8}
\end{figure}

\begin{figure}[tb]
  \centering
\includegraphics[width=0.98\linewidth]{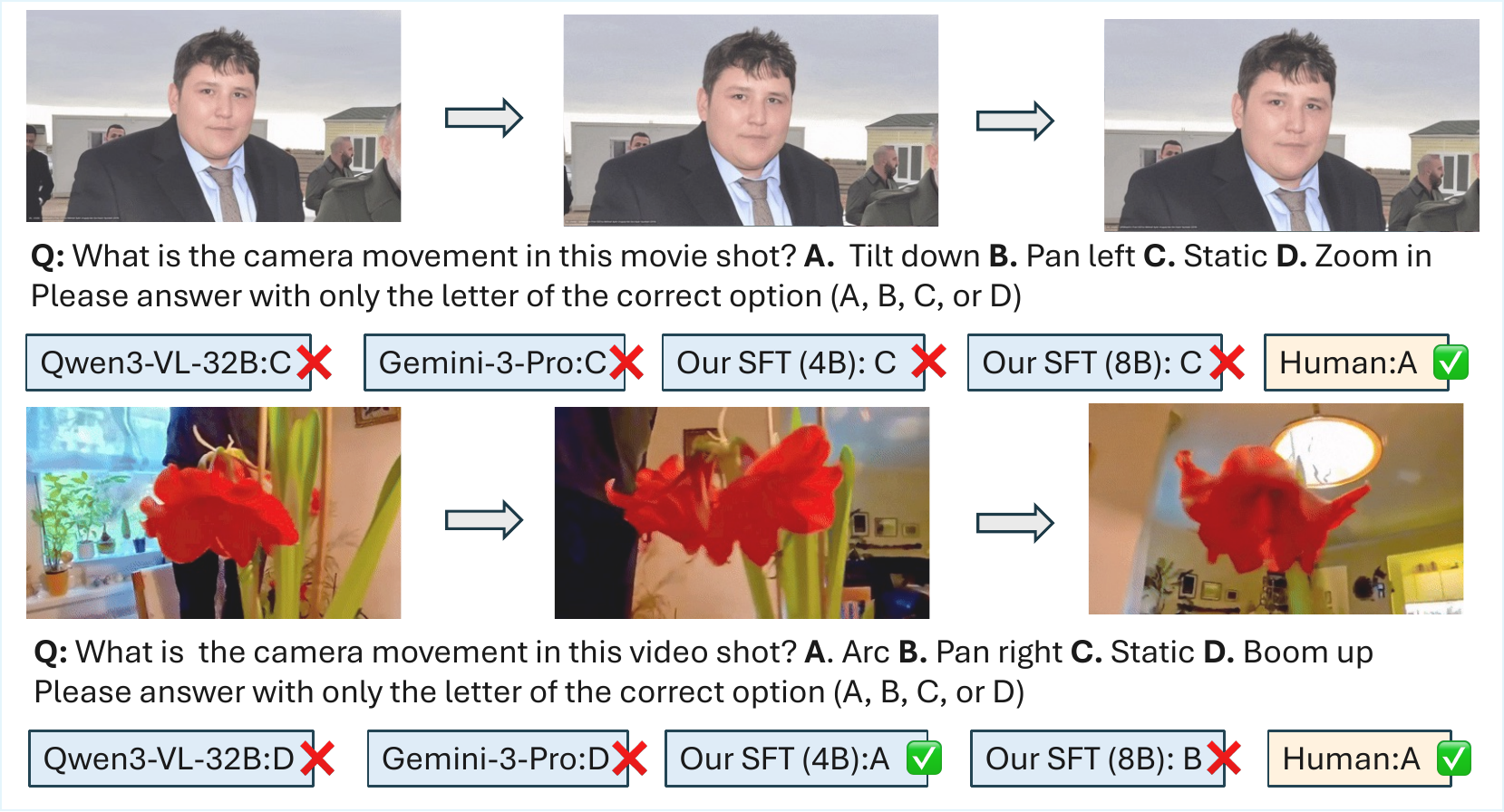}
  \caption{Qualitative comparison of model predictions under the MCQ setting.}
    \label{fig:supp_Veo_3_example_9}
\end{figure}

\noindent \textbf{Qualitative examples of model predictions.} We provide several qualitative examples in \cref{fig:supp_Veo_3_example_8} to illustrate the predictions of different models under the multiple-choice question (MCQ) setting. Compared with Gemini-3-Pro and Qwen3-VL-32B, our model produces more accurate predictions for camera movement. We also present several failure cases in \cref{fig:supp_Veo_3_example_9}.

\end{document}